\documentclass[lettersize,journal]{IEEEtran}
\usepackage{booktabs}
\usepackage{caption}
\usepackage{amsmath,amssymb,amsfonts}
\usepackage{mathrsfs}
\usepackage{algorithmicx}
\usepackage{algorithm}
\usepackage{adjustbox}
\usepackage{subfigure}
\usepackage{threeparttable}
\usepackage{longtable}
\usepackage{array}
\usepackage{hyperref}
\usepackage{enumitem}
\usepackage{textcomp}
\usepackage{stfloats}
\usepackage{soul}
\usepackage{multirow}
\usepackage{url}
\usepackage{verbatim}
\usepackage{graphicx}
\usepackage{cite}
\usepackage{algpseudocode}
\usepackage{lscape}
\usepackage{geometry}
\newcommand{\RNum}[1]{\uppercase\expandafter{\romannumeral #1\relax}}
\usepackage{bm}
\usepackage{tabularx}
\usepackage[justification=raggedright]{caption}
\graphicspath{{./fig/}}
\usepackage[labelformat=simple, labelsep=period]{caption}
\usepackage{xcolor}
\begin{document}
 \title{Graph-Driven Models for Gas Mixture Identification and Concentration Estimation on Heterogeneous Sensor Array Signals }
\author{Ding Wang, Lei Wang, Huilin Yin, Guoqing Gu, Zhiping Lin, \emph{Senior Member, IEEE},  Wenwen Zhang,  \emph{Member, IEEE}
\thanks{This work was partially supported by the National Natural Science Foundation of China under Grant 62203307. (\textit{Corresponding author: Wenwen Zhang.})  \par	
	Zhiping Lin and Wenwen Zhang are with School of Electrical and Electronic Engineering, Nanyang Technological University, Singapore 639798, Singapore (e-mail:  ezplin@ntu.edu.sg; wenwen.zhang@ntu.edu.sg  ).\par

	Ding Wang, Lei Wang and Huilin Yin are with College of Electronic and Information Engineering, Tongji University, Tongji University, Shanghai 201804, China (2310879@tongji.edu.cn; leiwang@tongji.edu.cn; yinhuilin@tongji.edu.cn).\par
     Guoqing Gu is with School of Civil Engineering, Yancheng Institute of Technology, Yancheng 224051, China (gqgu@ycit.edu.cn).
  \par		  
}}
\markboth{IEEE TRANSACTIONS ON INSTRUMENTATION AND MEASUREMENT}%
{Shell \MakeLowercase{\textit{et al.}}: A Sample Article Using IEEEtran.cls for IEEE Journals}


\maketitle

\begin{abstract}
Accurately identifying gas mixtures and estimating their concentrations are crucial across various industrial applications using gas sensor arrays. However, existing models face challenges in generalizing across heterogeneous datasets, which limits their scalability and practical applicability. To address this problem, this study develops two novel deep-learning models that integrate temporal graph structures for enhanced performance: a Graph-Enhanced Capsule Network (GraphCapsNet) employing dynamic routing for gas mixture classification and a Graph-Enhanced Attention Network (GraphANet) leveraging self-attention for concentration estimation. Both models were validated on datasets from the University of California, Irvine (UCI) Machine Learning Repository and a custom dataset, demonstrating superior performance in gas mixture identification and concentration estimation compared to recent models. In classification tasks, GraphCapsNet achieved over 98.00\% accuracy across multiple datasets, while in concentration estimation, GraphANet attained an R\textsuperscript{2} score exceeding 0.96 across various gas components. Both GraphCapsNet and GraphANet exhibited significantly higher accuracy and stability, positioning them as promising solutions for scalable gas analysis in industrial settings.
\end{abstract}


\section{Introduction}

\IEEEPARstart{I}{n} recent years, advancements in gas monitoring technologies have accelerated rapidly, driven by increasing industrial and environmental demands for accurate and real-time gas analysis. These advancements have found applications across a variety of critical industries, including ecological monitoring\cite{env1,env2}, healthcare\cite{hc1,hc2}, industrial safety\cite{ins1,ins2}, and chemical manufacturing\cite{chem1,chem2}. Traditional gas sensors, such as electrochemical sensors\cite{es}, infrared sensors\cite{is}, and metal-oxide sensors\cite{mos}, have long been employed for gas detection by exploiting the specific physical or chemical interactions between the sensor material and the target gas. 
However, differentiating multiple gases using traditional mathematical methods with a single sensor is challenging due to the similar interaction principles between various gases and sensor materials. 

This limitation necessitates the development of multi-sensor systems and advanced algorithms to distinguish gas mixtures more effectively. Integrating multiple sensors combined with machine learning, especially deep learning, algorithms offers a powerful approach to overcoming the limitations of traditional single-sensor systems, enabling more accurate analysis of complex gas mixtures. By learning patterns from multi-sensor data, these advanced systems can detect and differentiate gases more precisely, even in environments with overlapping chemical signatures.

Recent advances in gas identification have increasingly leveraged deep learning techniques to improve accuracy and robustness. For example, Pan \textit{et al.}\cite{lr1} introduced a multiscale convolutional neural network with attention (MCNA) for gas identification using semiconductor sensors, effectively extracting temporal features and dependencies between sensors. Their model achieves high accuracy with fewer parameters, making it suitable for embedded systems.
Wang \textit{et al.}\cite{lr2} proposed a Long short-term memory (LSTM) model with a self-attention mechanism to process gas sensor array data, achieving 99.6\% classification accuracy on complex gas mixtures. The attention mechanism performed better than traditional convolutional models in extracting temporal correlations in sensor data.
Dai \textit{et al.}\cite{lr3} developed a multitask learning-based model to classify gases and estimate their concentrations simultaneously. This approach enhanced generalization and robustness in different environmental conditions while reducing model complexity.
Luo \textit{et al.}\cite{lr4} proposed an electronic nose system based on the Fourier series for gas identification and concentration estimation. This method improved the interpretability of sensor signals and maintained high classification accuracy across various gas mixtures. However, much of the current research in gas mixture identification has concentrated on developing deep-learning models specifically designed for individual datasets. While these models achieve high accuracy within their respective datasets, they often require significant adjustments, including hyperparameters re-selection, when applied to new datasets. This lack of generalizability poses a significant limitation, especially in industrial applications where large-scale and rapid deployment of models across diverse conditions is crucial. The necessity for model retraining and fine-tuning with each new dataset not only increases the time and computational resources required but also limits the scalability and adaptability of these models. To address these challenges, our research focuses on developing robust deep-learning models capable of generalizing across heterogeneous datasets, thereby fulfilling the growing industrial demand for efficient and versatile gas mixture identification systems.

This study introduces a Graph-Enhanced Capsule Network (GraphCapsNet) for gas mixture identification and a Graph-Enhanced Attention Network (GraphANet) for estimating the concentrations of individual components within a gas mixture. Our approach begins by representing sensor signals as a temporal graph, capturing a more nuanced data structure. Subsequently, a Graph Convolutional Network (GCN) is employed to extract relevant features from these signals. In GraphCapsNet, we utilize a dynamic routing mechanism \cite{dynamic}, while in GraphANet, we integrate a self-attention mechanism \cite{sa} to analyze the extracted features within the graph framework, enhancing both identification and concentration estimation accuracy. To assess the model's robustness with heterogeneous data, we conducted extensive experiments on both a UCI Machine Learning Repository dataset (UCI dataset) \cite{uci} and a custom dataset. The results demonstrate that our model not only outperforms existing methods in overall accuracy for gas mixture identification but also achieves superior concentration estimation, highting its potential for broad industrial applications.
The significant contributions of this study are as follows:

1) 
We developed two novel deep-learning models for transforming sensor signals: GraphCapsNet and GraphANet. These models were rigorously validated on two datasets with distinct data structures using identical hyperparameters. Both models achieved classification accuracies exceeding 98\%, with R\textsuperscript{2} values consistently above 0.96 for concentration estimation across all datasets. These results demonstrate exceptional performance and robust generalization, setting a new benchmark in sensor data analysis.

2) We demonstrated that the models are highly adaptable and robust when applied to heterogeneous datasets with varying structures, collection conditions, and data types. Despite these differences, the models consistently delivered high performance, minimizing the need for customization and enhancing their practicality and scalability in diverse sensor data environments. Furthermore, to address variations in recognition difficulty across gas-related datasets, we introduced a cross-dataset weighted accuracy metric for gas identification, providing an effective measure of model performance across multiple datasets.

3) 
Our models process raw sensor signals directly, eliminating the need for complex preprocessing steps such as filtering or normalization. This streamlined workflow reduces computational overhead and operational costs compared to models requiring extensive preprocessing. By preserving data integrity, this approach enables more reliable and efficient gas composition estimation, addressing industry demands for real-time monitoring and rapid decision-making while maintaining high accuracy.


The remainder of this article is organized as follows: Section \uppercase\expandafter{\romannumeral2} introduces the gas sensor array signal acquisition system and the data collection process. Section \uppercase\expandafter{\romannumeral3} details the architectures of GraphCapsNet and GraphANet. Section \uppercase\expandafter{\romannumeral4} outlines the experimental settings, while Section \uppercase\expandafter{\romannumeral5} presents the experimental results and their analysis. Finally, the conclusions are summarized in Section \uppercase\expandafter{\romannumeral6}.

\section{Gas Sensor Array Signal Acquisition System and Data Collection}
\subsection{Gas sensor array and signal acquisition}
\subsubsection{The overview of sensor array from UCI dataset and our lab}
\begin{figure}[htbp]
	\centering
	\subfigure[]{ 
		\begin{minipage}[t]{0.5\linewidth}
			\centering
			\includegraphics[width=4.4cm]{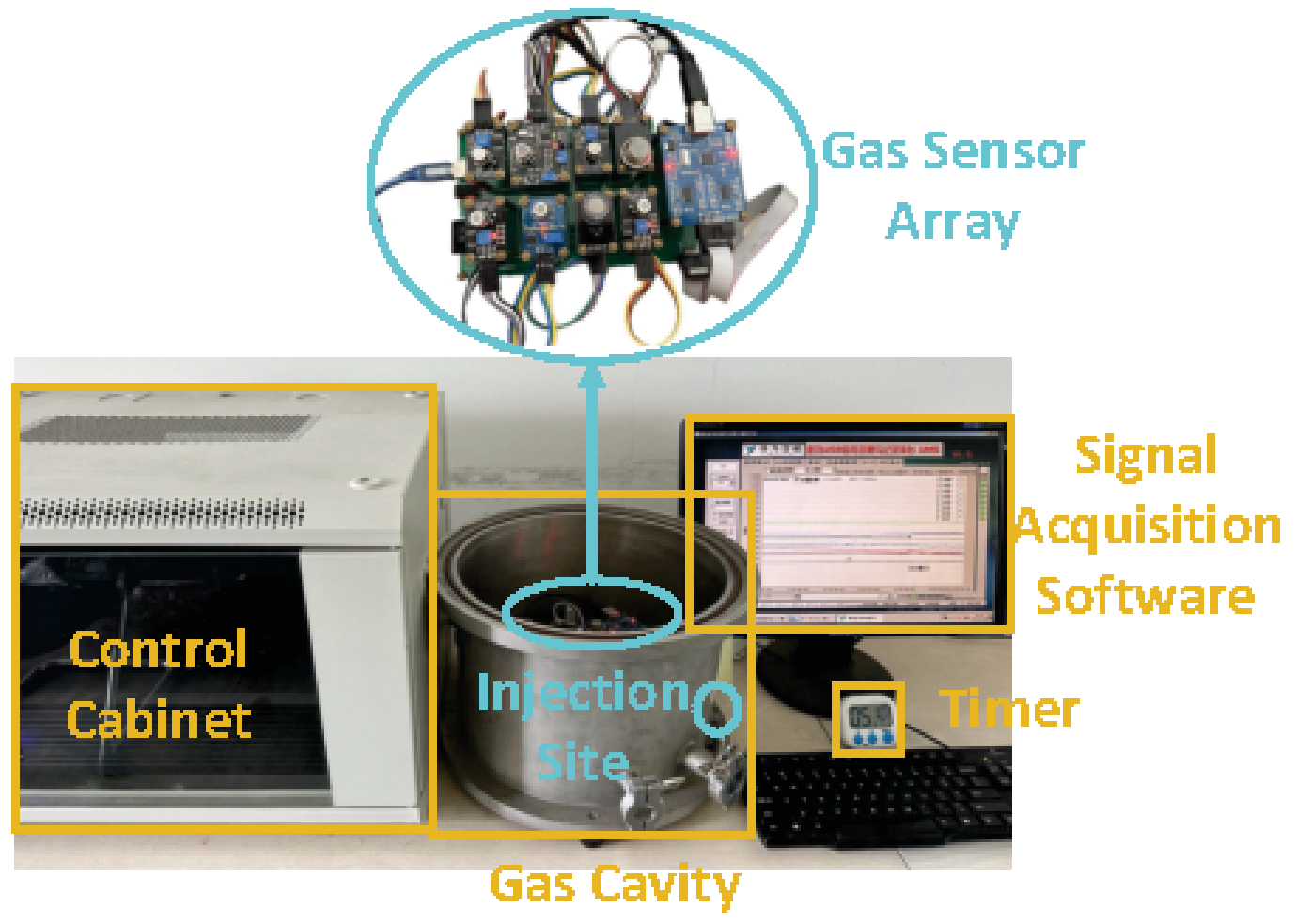}
		\end{minipage}%
	}%
 
	\subfigure[]{ 
		\begin{minipage}[t]{0.5\linewidth}
			\centering
			\includegraphics[width=4.4cm]{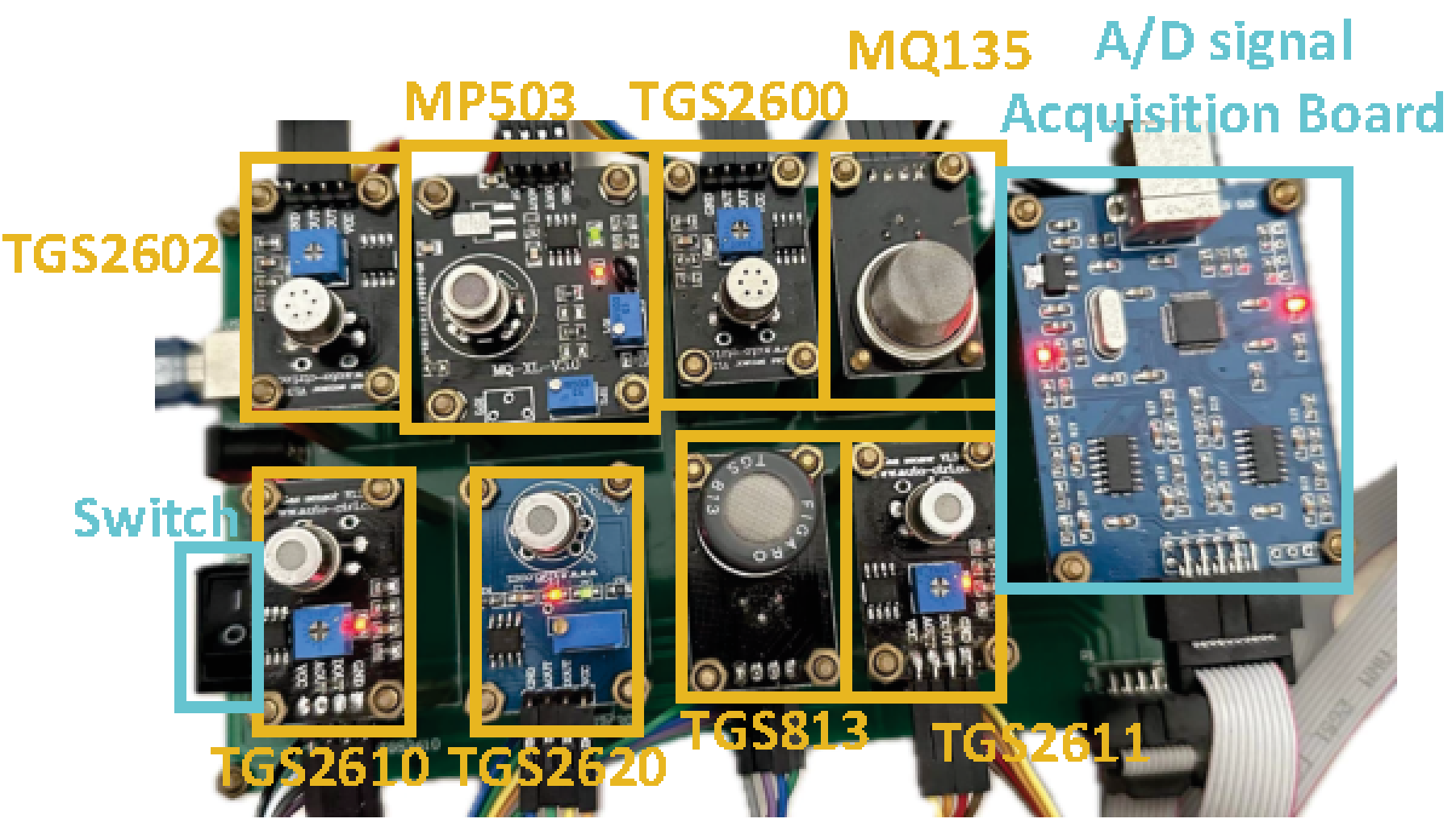}
		\end{minipage}%
	}%
	\caption{(a) Standard gas configuration system. (b) Custom-developed gas sensor array.}
	\label{hardware}
\end{figure}

The experimental setups in this study involve two distinct data collection systems: one from an open-source dataset in the UCI repository \cite{uci}, which employs 16 sensors housed in a 60 mL chamber, and the other from our laboratory, shown in Fig. \ref{hardware}, which employs eight different sensors inside a larger 5-liter metal chamber. In our setup, gas samples are introduced by calculating the desired concentration and injecting the gas through an injection port using a syringe. The Table \ref{hh} summarizes critical differences between the two systems, including sensor count, chamber size, gas injection method, and control mechanism.

\begin{table}[h!]
\centering
\caption{Comparison between the UCI and our custom sensor array system.}
\begin{tabular}{@{}lcc@{}}
\toprule
\textbf{Feature}             & \textbf{UCI hardware}              & \textbf{Our Lab's hardware}                   \\ \midrule
\textbf{Number of sensors}   & 16                       & 8                             \\ 
\multirow{4}{*}{\textbf{Sensor types}}        & TGS2600  *4                         & MQ135, TGS813          \\
                             & TGS2602 *4                          & TGS2602, TGS2610    \\
                             & TGS2610 *4                          & TGS2611, TGS2620        \\ 
                             & TGS2620 *4                         & TGS2602, MP503                   \\ 
\textbf{Chamber size}        & 60 mL                              & 5 L                                \\ 
\textbf{Control method}      & Automated                  & Manual Control   \\
\textbf{Gas injection}       & 300 mL/min     & Syringe        \\ 
\textbf{Sampling rate}  & 100Hz                     & 10Hz  \\ 
 \bottomrule
\end{tabular}
\label{hh}
\end{table}

\subsubsection{Data acquisition protocol}
Our system's systematic data collection follows a well-defined sequence to ensure accurate and repeatable measurements. It consists of four key steps:

\begin{enumerate}[label=(\roman*)]
    \item \textbf{Calculation of target gas volume}: To determine the volume \(V_{in}\) of gas to be injected into the chamber, we calculate it based on the desired concentration of the target gas \(C_{t}\). The following formula is employed for this calculation:
    \[
    V_{in} = \frac{C_{t} \times V_{cav}}{\beta} \times 10^{-6}
    \]
    where \(\beta\) denotes the proportional coefficient of the target gas in the gas cylinder, and \(V_{cav}\) is the volume of the chamber, which is 5 L in our case.
    
    \item \textbf{Target gas injection}: After the sensor signals have stabilized, the target gas is extracted from the cylinder using a pressure-reducing valve and transferred into a gas collection bag. A syringe is then employed to draw the calculated volume of gas from step 1, which is injected into the chamber through an injection port to ensure the correct concentration of the target gas for each experiment.
    
    \item \textbf{Dynamic response signal acquisition}: Following the gas injection, gas sensor array is activated via the LabVIEW interface to record the sensors' dynamic response. This process continues for approximately 200 seconds, during which the system continuously monitors and captures the sensor signals as the gas interacts with the sensor array.
    
    \item \textbf{Cavity environment cleaning}: After the recording, an air pump removes the remaining gas from the chamber. This process continues until the sensor array's readings return to baseline, indicating that the chamber has been refilled with clean air and is ready for the next experiment.
\end{enumerate}

Steps \expandafter{\romannumeral2} to \expandafter{\romannumeral4} are repeated for each gas sample until all experimental data have been collected, ensuring a comprehensive dataset.

\begin{table*}[htbp]
\centering
\caption{Dataset information including gas type, sample number, concentration range, and data splitting conditions for concentration estimation and gas classification tasks.}
\begin{threeparttable}
\begin{tabular}{@{}llccccccc@{}}
\toprule
\multicolumn{2}{c}{\multirow{2}{*}{Dataset}} & \multirow{2}{*}{Gas type} & \multirow{2}{*}{Sample size} & \multirow{2}{*}{\shortstack{Concentration range\\(ppm)}} & \multicolumn{2}{c}{Concentration estimation} & \multicolumn{2}{c}{Gas classification}\\ \cmidrule(l){6-7} \cmidrule(l){8-9} 
 &  &  &  &  & Train-val set & Test set & Train-val set & Test set \\ \midrule
\multirow{10}{*}{UCI} & \multirow{5}{*}{\shortstack{Group\\A}}& Air & 89 & -- & 74 & 15 & 74 & 15 \\
 &  & CO & 71 & 200-533.33 & 59 & 12 & 59 & 12 \\
 &  & C\textsubscript{2}H\textsubscript{4} & 111 & 6.67-20 & 93 & 18 & 93 & 18\\
 &  & Mixture & 100 & CO:200-520 C\textsubscript{2}H\textsubscript{4}:6.67-20 & 84 & 16 & 84 & 16 \\ 
 &  & \textbf{Total} & \textbf{371} & -- & \textbf{310} & \textbf{61}  &  &\\ \cmidrule(l){2-7}
 & \multirow{5}{*}{\shortstack{Group\\B}} & Air & 98 & -- & 82 & 16 & 82 & 16\\
 &  & CH\textsubscript{4} & 76 & 66.66-296.67 & 63 & 13 & 63 & 13\\
 &  & C\textsubscript{2}H\textsubscript{4} & 89 & 6.67-20 & 74 & 15 & 74 & 15 \\
 &  & Mixture & 86 & CH\textsubscript{4}:66.66-220 C\textsubscript{2}H\textsubscript{4}:6.67-18.33 & 72 & 14 & 72 & 14\\ 
 &  & \textbf{Total} & \textbf{349} & -- & \textbf{291} & \textbf{58} & \textbf{601} & \textbf{119}\\ \midrule
\multicolumn{2}{c}{\multirow{4}{*}{\shortstack{Custom\\Dataset}}} & H\textsubscript{2} & 150 & 10--1000 & 126 & 24 & 126 & 24\\
 &  & C\textsubscript{2}H\textsubscript{4} & 150 & 10--1000 & 126 & 24 & 126 & 24\\
 &  & Mixture & 300 & H\textsubscript{2}:10-980 C\textsubscript{2}H\textsubscript{4}:30-1000 & 252 & 48 & 252 & 48\\ 
 &  & \textbf{Total} & \textbf{600} & -- & \textbf{504} & \textbf{96} & \textbf{504} & \textbf{96}\\ \bottomrule
\end{tabular}
\end{threeparttable}
\label{data}
\end{table*}
\subsection{ UCI and self-developed sensor array  datasets}

\begin{figure}[htbp]
	\centering
	\subfigure[]{ 
		\begin{minipage}[t]{0.9\linewidth}
			\centering
			\includegraphics[width=\columnwidth]{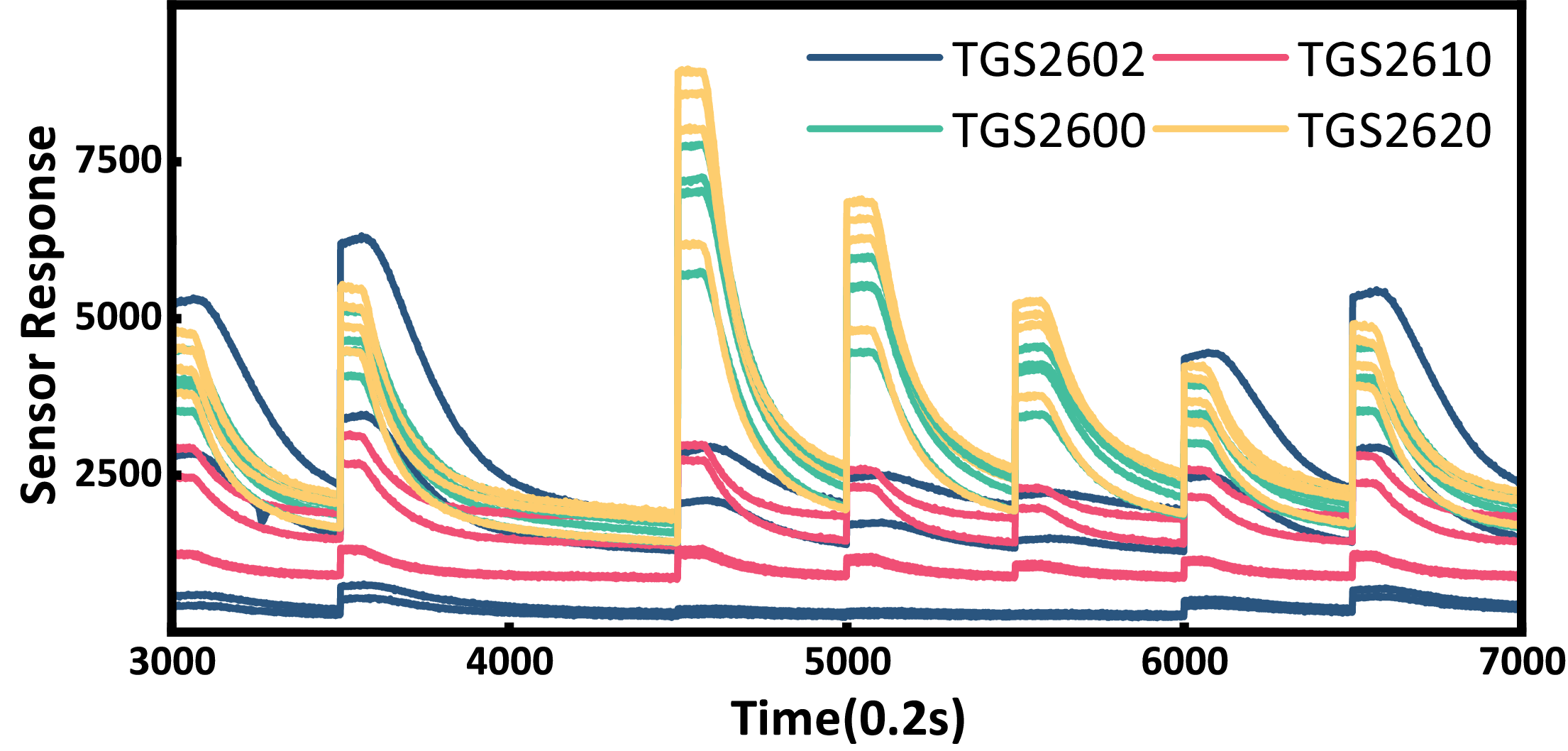}
		\end{minipage}%
	}%
 
	\subfigure[]{ 
		\begin{minipage}[t]{0.9\linewidth}
			\centering
			\includegraphics[width=0.7\columnwidth]{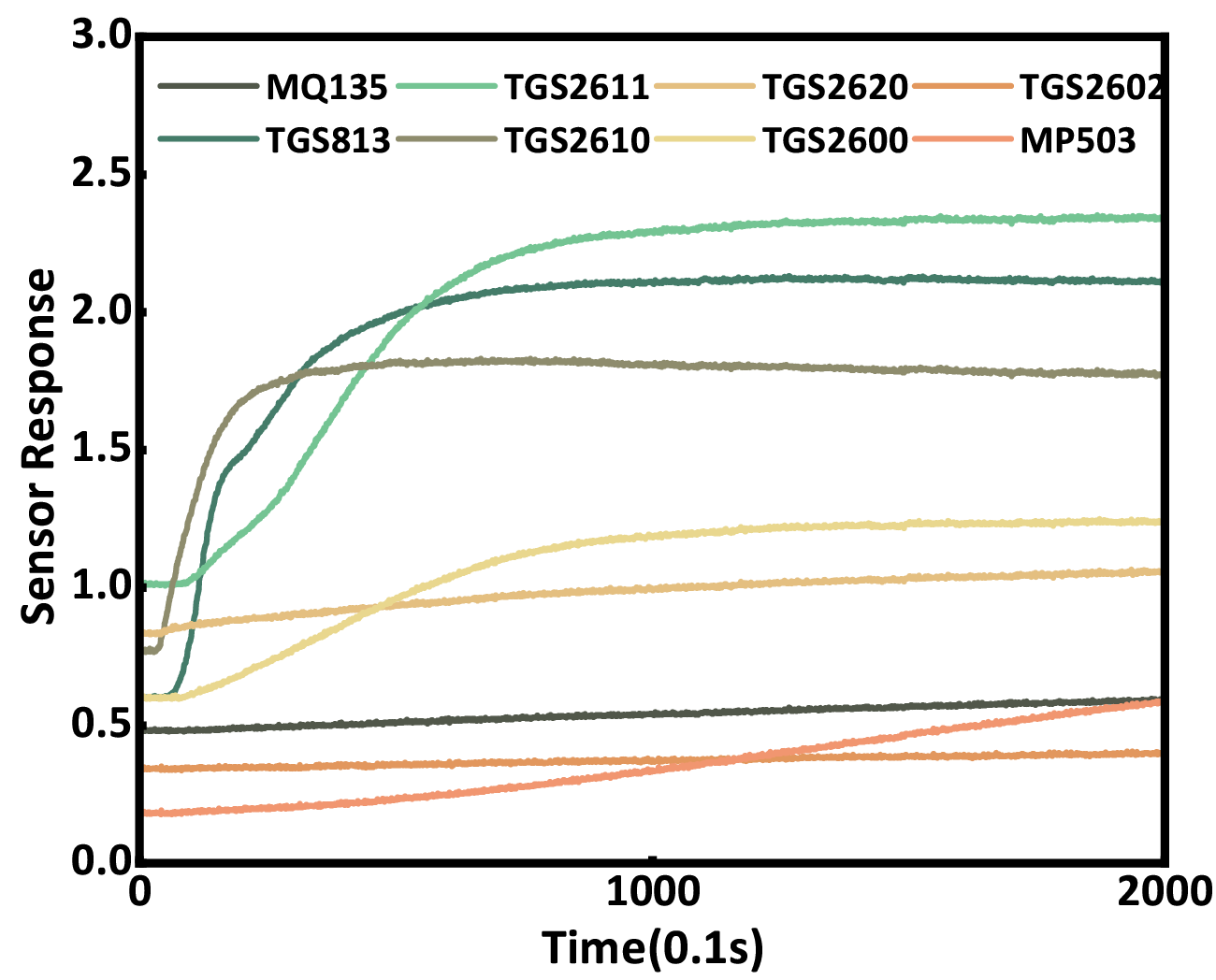}
		\end{minipage}%
	}%
	\caption{(a) Data segment of UCI dataset. (b) Custom data sample.}
	\label{sample}
\end{figure}  

\subsubsection{UCI sensor array dataset}
\begin{figure}[htbp]
	\subfigure[CO-C\textsubscript{2}H\textsubscript{4}]
	{\begin{minipage}[c]{0.32\columnwidth}
      \centering
      \includegraphics[width=\columnwidth]{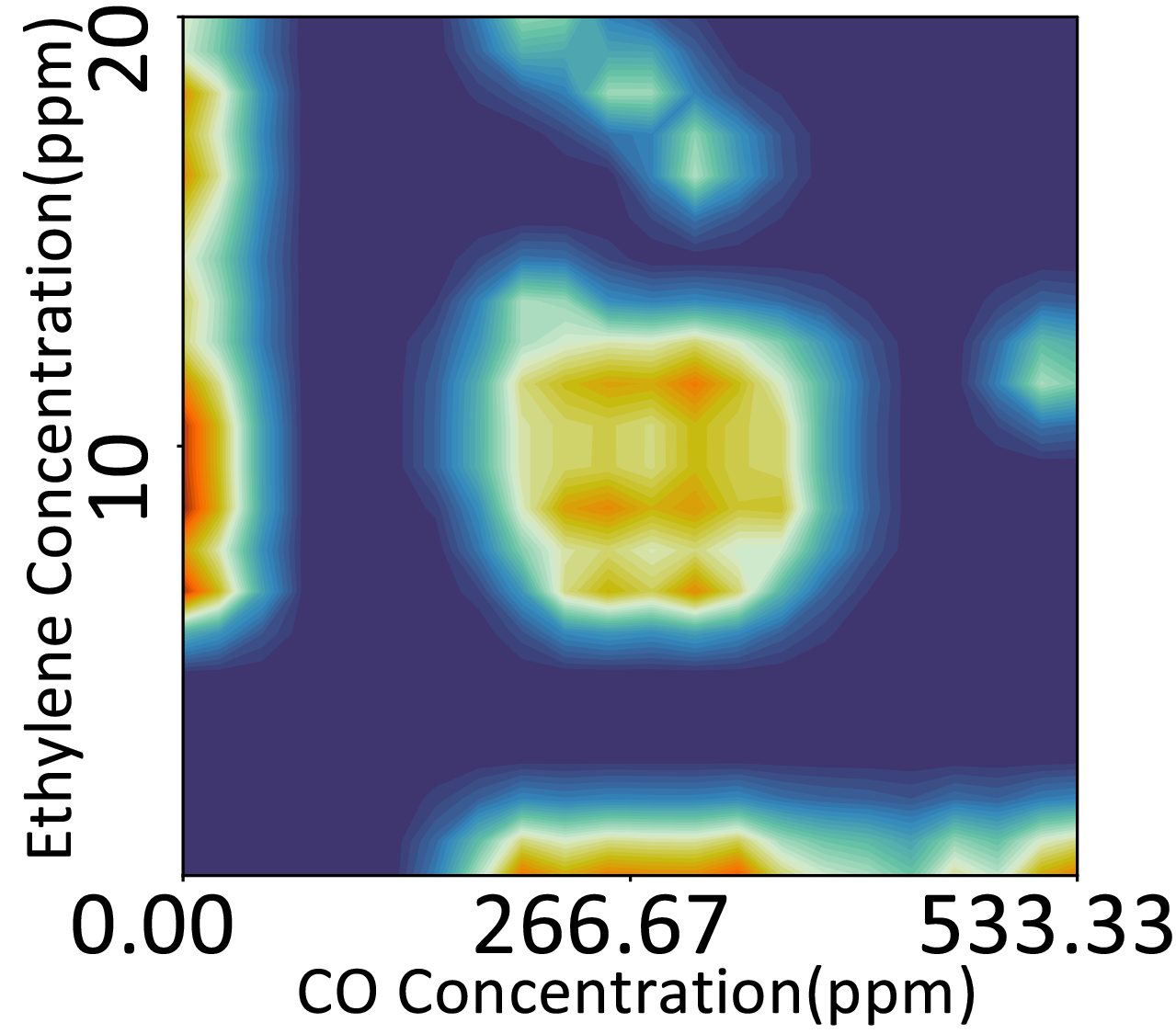}
      \end{minipage}}
	\subfigure[CH\textsubscript{4}-C\textsubscript{2}H\textsubscript{4}]
	{\begin{minipage}[c]{0.32\columnwidth}
	 \centering
      \includegraphics[width=\columnwidth]{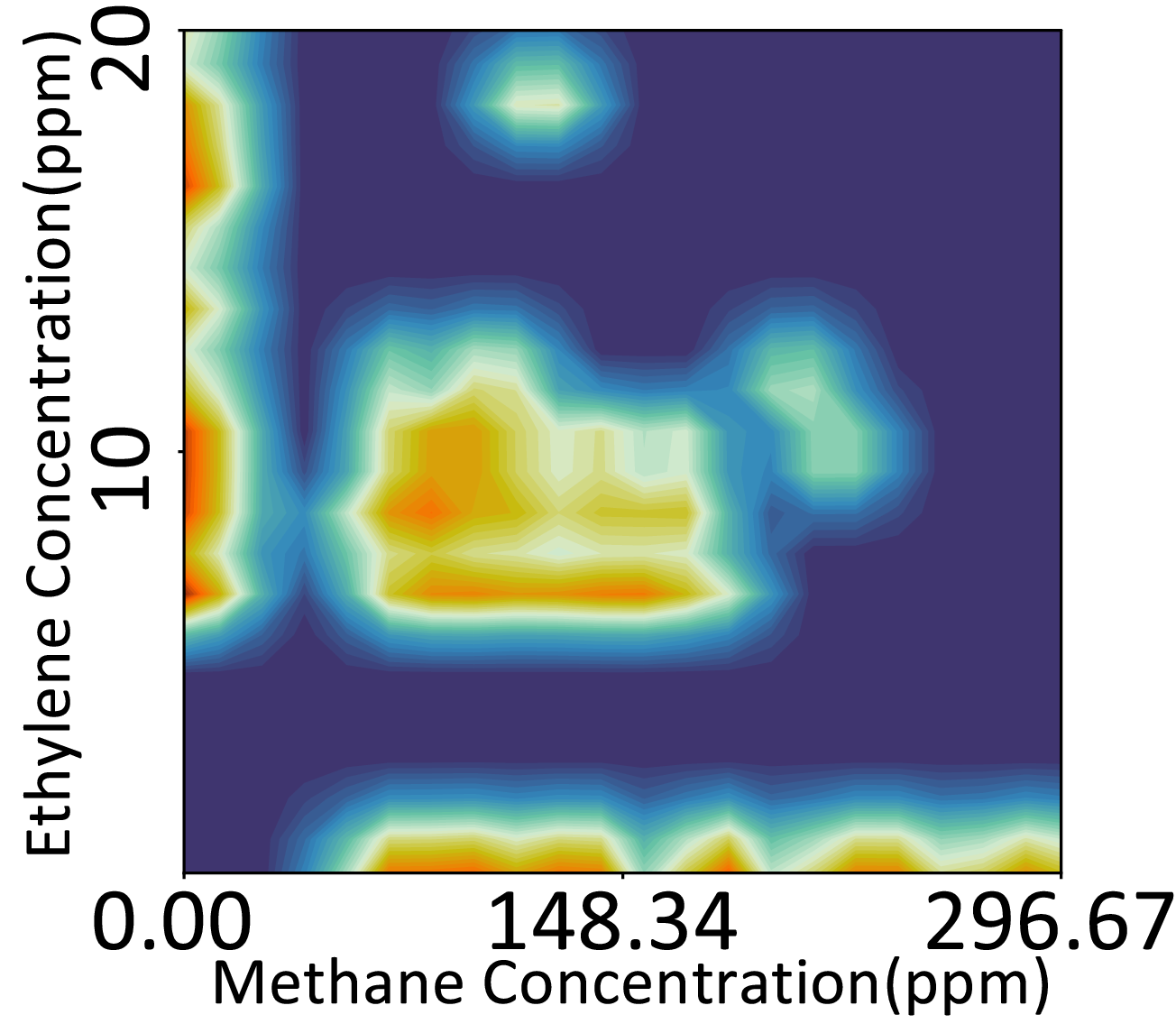}
      \end{minipage}}
	\subfigure[H\textsubscript{2}-C\textsubscript{2}H\textsubscript{4}]
      {\begin{minipage}[c]{0.31\columnwidth}
	 \centering
      \includegraphics[width=\columnwidth]{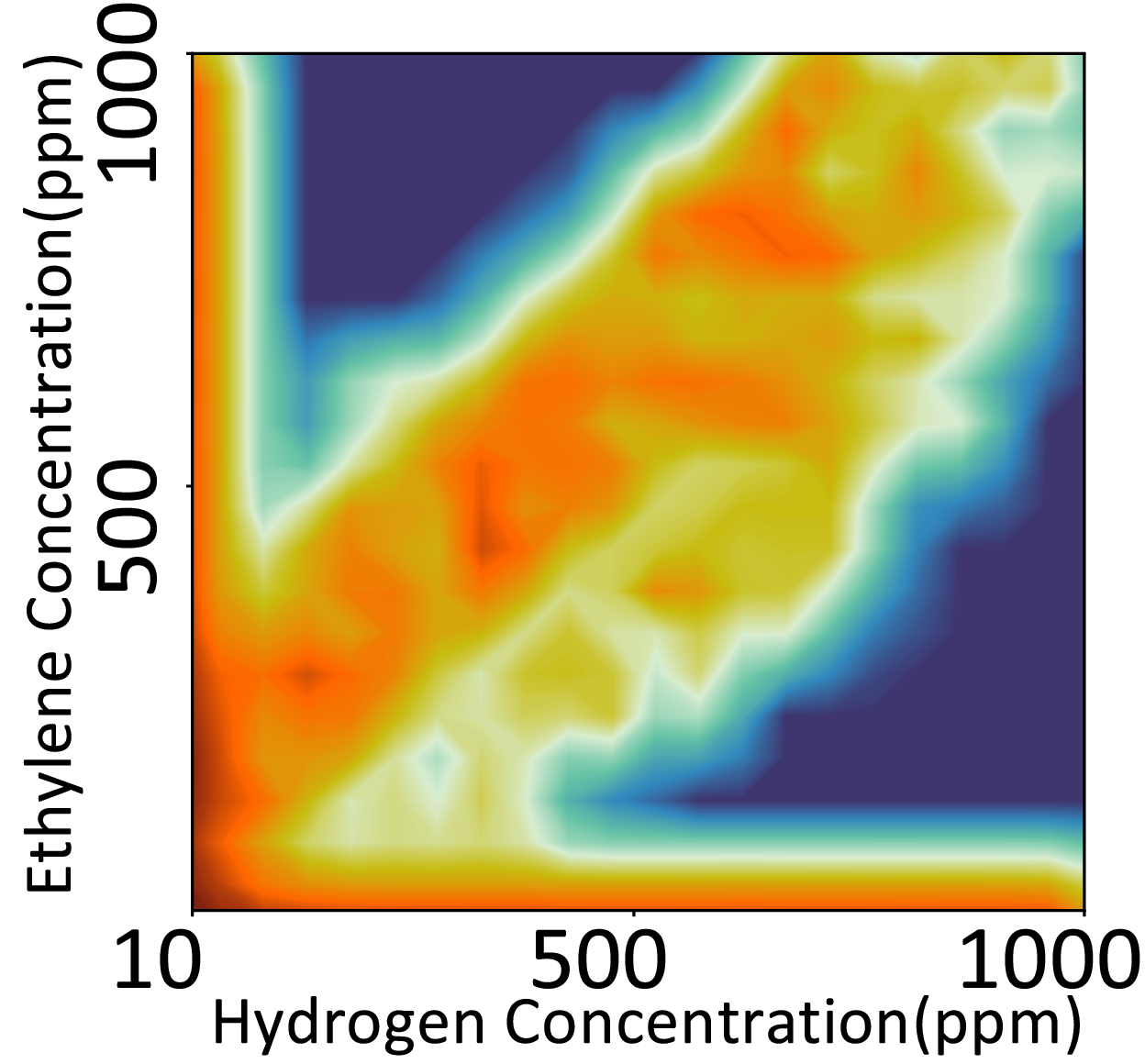}
    \end{minipage}}
    \caption{The heat map of data distribution. (a) CO-C\textsubscript{2}H\textsubscript{4} from UCI dataset (b) CH\textsubscript{4}-C\textsubscript{2}H\textsubscript{4} from UCI dataset (c) H\textsubscript{2}-C\textsubscript{2}H\textsubscript{4} from custom dataset.}
\label{ddhm}
\end{figure}

UCI dataset contains two groups: C\textsubscript{2}H\textsubscript{4}-CO (group A) and C\textsubscript{2}H\textsubscript{4}-CH\textsubscript{4}  (group B). The C\textsubscript{2}H\textsubscript{4}, CO, and CH\textsubscript{4} concentrations range from 0 to 20 ppm, 0 to 533.33 ppm, and 0 to 296.67 ppm, respectively. The concentration distribution heat map can be found in Fig. \ref{ddhm} (a) and Fig. \ref{ddhm} (b).  The sensor data are continuously recorded for 12 hours at a sampling frequency of 100 Hz. There are 8387665 total time points. This dataset is downsampled by 5 Hz to remove the redundant data. The downsampled data segment can be found in Fig. \ref{sample} (a). In different environments, sensors' responses are different. In addition, the initial 100 s of the heating phase and the last 100 s of the air phase were skipped. The detailed information on groups A and B of the UCI dataset, including gas type, sample size, and concentration range, is presented in Table \ref{data}. The data processing of the UCI data can be found in Fig. \ref{dp}.  The data was then divided into graphs of different forms depending on the composition and concentration of the gas. Each time point can be considered a node in the graph, and the sensing data acquired by each sensor at each instant can be regarded as its attribute. There are sixteen sensors, each corresponding to one of the sixteen features associated with each node. 
The longest graph in the dataset contains 4351 nodes, corresponding to sensor data spanning 14 minutes, while the shortest graph contains only five nodes, corresponding to sensor data spanning one second.

\subsubsection{self-developed sensor array dataset}
\begin{figure}
\centering
{\begin{minipage}[c]{\columnwidth}
\includegraphics[width=\columnwidth]{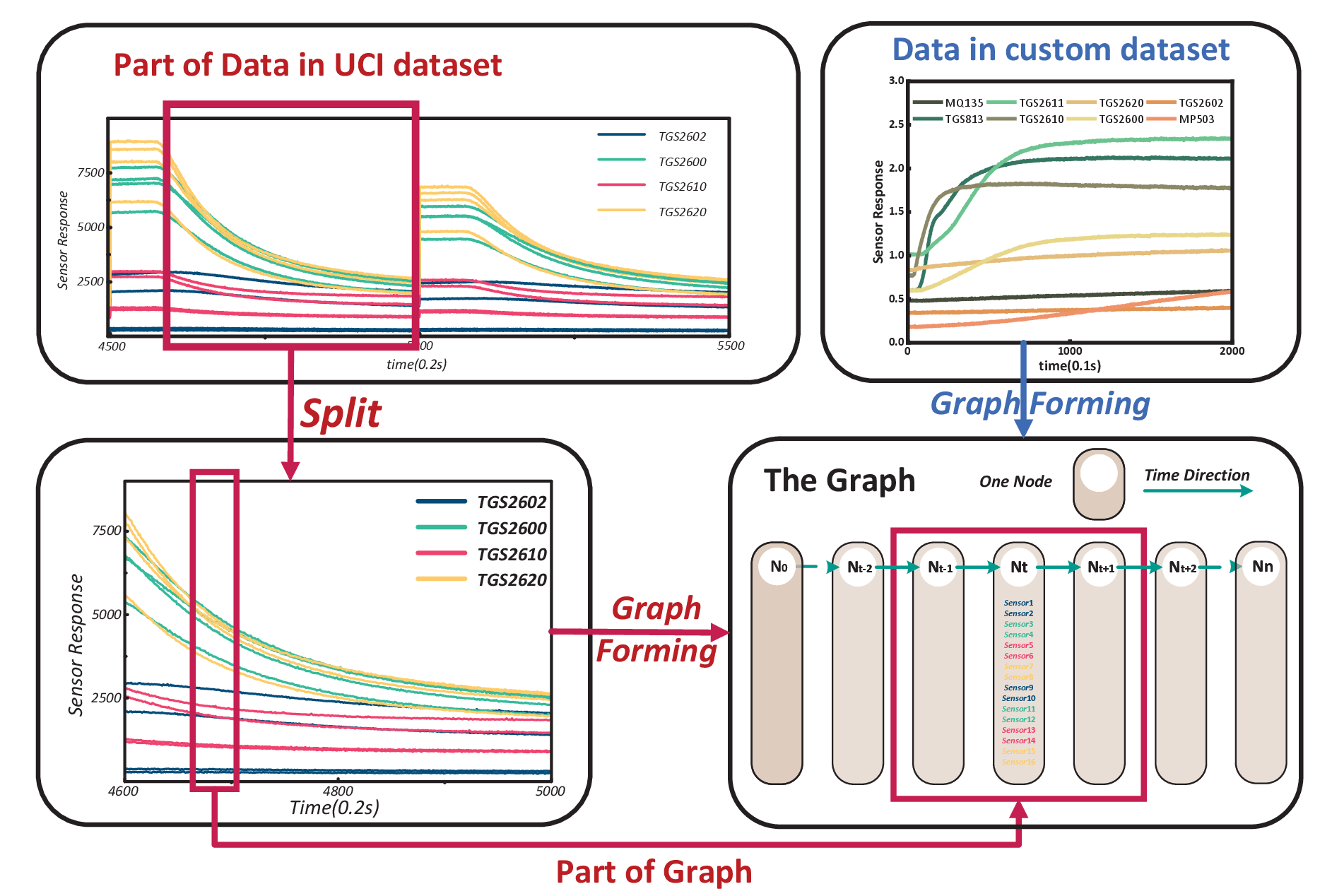}
\end{minipage}}
\caption{The data processing of UCI data and custom data.}
\label{dp}
\end{figure}

We developed an eight-sensor array to collect signals for H\textsubscript{2}, C\textsubscript{2}H\textsubscript{4}, and their mixtures. A sample of custom dataset could be found in Fig. \ref{sample} (b). The sampling frequency was set to 10 Hz, with a collection duration of 200 seconds. Six hundred samples were collected, including 150 samples of pure H\textsubscript{2}, 150 samples of pure C\textsubscript{2}H\textsubscript{4}, and 300 samples of gas mixtures. The concentrations of H\textsubscript{2} and C\textsubscript{2}H\textsubscript{4} were uniformly distributed between 0 and 1000 ppm. The detailed information on the dataset collected from our self-developed sensor array, including gas type, sample size, and concentration range, is presented in Table \ref{data}. The concentration distribution heat map can be found in Fig. \ref{ddhm} (c). Compared to the UCI dataset, processing the custom data is more straightforward, as each sample is collected independently, requiring only the same graph-based transformation method applied to the UCI data, as seen in Fig. \ref{dp}. Each time point is treated as a node in the graph, and the sensor readings at each moment are considered node attributes. The main distinction in the custom dataset lies in the number of features per node, with each node containing eight attributes corresponding to the eight sensors.
\section{Methodology}
\subsection{The architecture of GraphCapsNet}
\begin{figure*}
   \centering
   \includegraphics[width=0.95\textwidth]{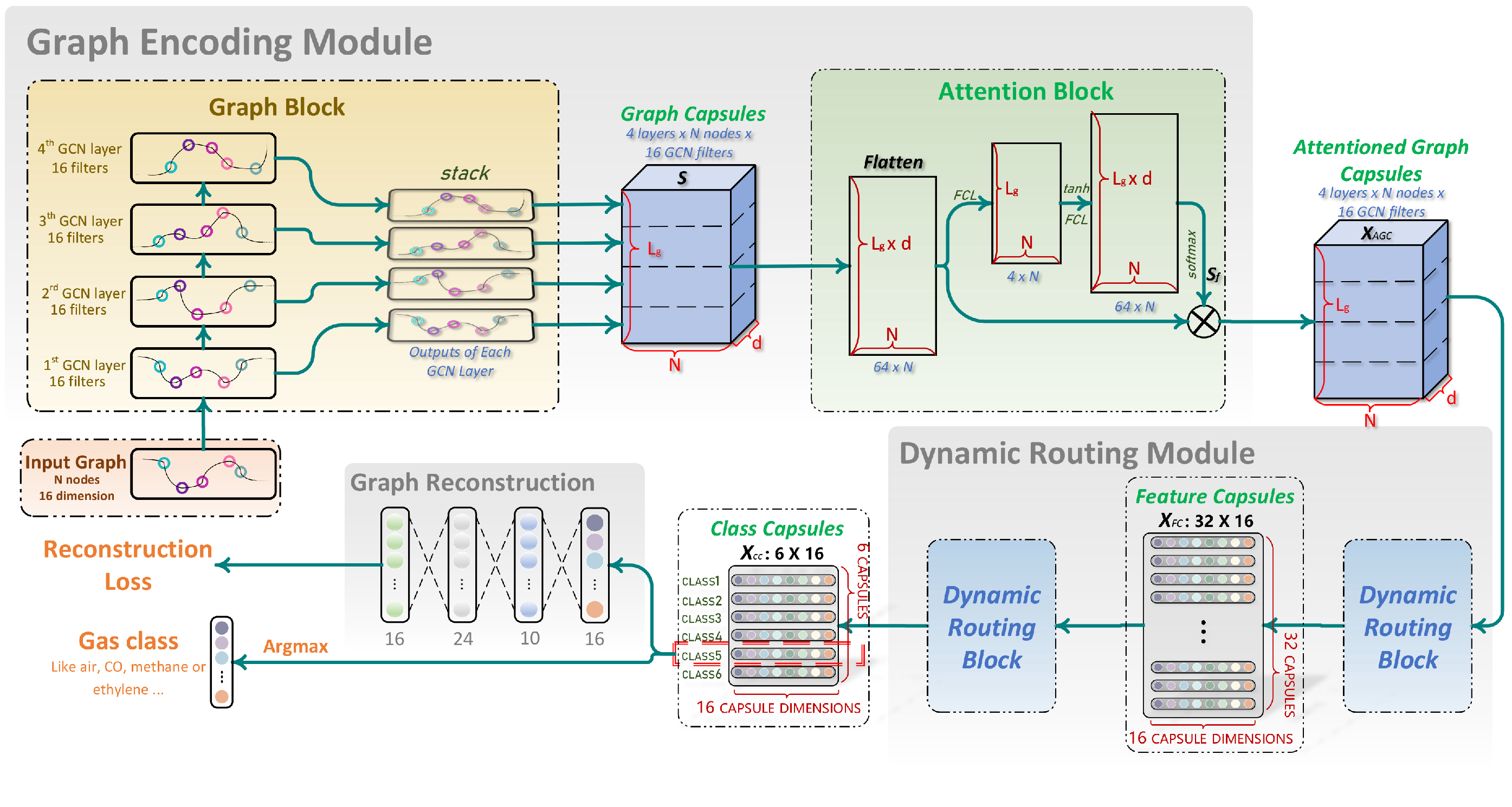}
   \caption{Architecture of the GraphCapsNet model for gas recognition.}
   \label{modelc}
\end{figure*}

As illustrated in Fig. \ref{modelc}, GraphCapsNet comprises three main modules: the graph encoding module, the dynamic routing module, and the graph reconstruction module. The graph encoding module consists of two blocks: the graph block and the attention block. The sensor data is represented as a chain graph before being processed by the graph block, which utilizes a Graph Convolutional Network (GCN) \cite{gcn}. The outputs from various layers of the GCN are stacked together. An attention block is then applied to further refine the graph by adjusting weights based on the learned attention values. The dynamic routing module includes two dynamic routing blocks. The first block routes the attended graph capsules to feature capsules, which are subsequently routed dynamically to class capsules. The graph reconstruction module is responsible for calculating the reconstruction loss, which is then weighted and combined with the classification loss to yield the final loss. 

    


\subsubsection{Graph encoding module}
The core of the graph encoding module consists of four GCN layers. Specifically designed for graph-structured data, GCNs are widely employed in temporal studies \cite{gt}\cite{gs} for their ability to effectively capture spatiotemporal dependencies. By integrating graph structures with temporal information, they significantly enhance prediction accuracy. In this study, we transform the array sensor data $\textbf{D} \in \mathbb{R}^{N \times F}$ into a unidirectional chain $\mathscr{K} = \{ \{i\} \mid i = 1, 2, \ldots, T \} \cup \{ \{i, i+1\} \mid i = 1, 2, \ldots, T-1 \}$
, giving $\textbf{D}$ a temporal topological structure, which is also could be expressed as $\mathcal{G}=(\textbf{X}, \textbf{A})$. $\textbf{X}$ denotes the feature matrix, while $\textbf{A}$ denotes the adjacency matrix, the two components of the input for the GCN block. After passing through four GCN layers, the outputs from each layer are stacked to form Graph Capsules $\textbf{S} \in \mathbb{R}^{L_{g} \times N \times d}$, where $L_g$ is the number of GCN layers, $N$ is the number of nodes, and $d$ denotes the number of GCN filters. Afterward, the attention module processes the $\textbf{S}$ by flattening it and passing it through two fully connected layers to obtain the attention value, then multiplied by $\textbf{S}_{f}$ and reshaped into attentioned graph capsules $\textbf{X}_{AGC}$. Graph encoding module has parameters of $L_{g}$ = 4, $d$ = 8. \par
\subsubsection{Dynamic routing module}
Dynamic routing, designed to improve the representation of spatial hierarchies and object transformations\cite{dynamic}, involves updating routing coefficients to maximize agreement between low-level and high-level capsules' outputs. This method enhances the network's ability to capture complex patterns and relationships in the data.

This study employs the dynamic routing mechanism twice in the dynamic routing module. The first instance is employed to process attentioned graph capsules $\textbf{X}_{AGC}$, which have been processed by the graph encoding module, into feature capsules $\textbf{X}_{FC}\in \mathbb{R}^{P \times d'}$, where $P$ is the total number of feature capsules and $d'$ is the dimensionality of each feature capsule. This step involves embedding the feature information from different GCN layers, establishing connections and weight allocations, and eliminating redundant information to form the feature capsules. The second instance of dynamic routing further processes the feature capsules $\textbf{X}_{FC}$, capturing richer and more meaningful graph structural information, ultimately creating the final class capsules $\textbf{X}_{CC}\in \mathbb{R}^{C \times d'}$, where $C$ represents the number of gas types. The argmax function determines the final result. Dynamic Routing Module has parameters of $P$ = 32, $d'$ = 16. 

\subsubsection{The loss}
The loss is the weighted sum of classification and reconstruction losses\cite{dynamic}.

\begin{equation}
Loss=Loss_c + \theta Loss_r
\end{equation}
where $Loss_c$ and $Loss_r$ denote classification loss and reconstruction loss, respectively. $\theta$ is set to 0.0005.
The following formula can be used to calculate the classification loss $Loss_c$:
\begin{equation}
\resizebox{0.99\columnwidth}{!}{$
Loss_c=\sum_k \{ {T_kmax(0,m^+- \Vert{\textbf{c}_k}\Vert)^2+\lambda(1-T_k)max(0,\Vert{\textbf{c}_k}\Vert - m^- )^2}\}
$}
\end{equation}
where the modulus length of the capsule output vector is denoted by $\textbf {c}_k$. This value denotes the magnitude of the class probability. The variable $\lambda$ is the down-weight of the loss for the non-existent object class. The values $m^{+}$ and $m^{-}$ denote the set penalty term thresholds. If the input graph belongs to class $k$, then the value of $T_k$ is 1. To prevent initial learning from shortening all class capsules, $\lambda$ is used, especially when $k$ is high. In this work, the values of $m^{+}$ and $m^{-}$ are set to 0.9 and 0.1, respectively, and $\lambda$ is set to 0.5. 

The reconstruction loss is based on the discrepancy between the reconstruction and original graphs. The reconstruction validates the model's ability to learn and accurately represent input features by transforming high-level capsule outputs into original images. This process enhances the model's robustness against variations in input data, such as transformations and distortions, ultimately improving its performance in complex visual tasks. The following formula is employed to determine the system reconstruction loss:
\begin{equation}
Loss_r= \frac{\sum_i{P_i(\textbf{d}_i-\textbf{m}_i)^2}}{\sum_i{P_i}}+ \frac{\sum_i{(1-P_i)(\textbf{d}_i-\textbf{m}_i)^2}}{\sum_i{1-P_i}}
\end{equation}
where $\textbf{m}_i$ denotes the number of nodes with the attribute $i$ appearing in the input graph, $\textbf{d}_i$ denotes the corresponding decoded value. If there are nodes in the input graph with the attribute $i$,  $P_i$ = 1. The equation is employed to avoid reducing reconstruction loss by setting all decoded values to 0, particularly when most of the ground truth are 0.

\subsection{The achitecture of GraphANet}
\begin{figure*}
   \centering
   \includegraphics[width=0.95\textwidth]{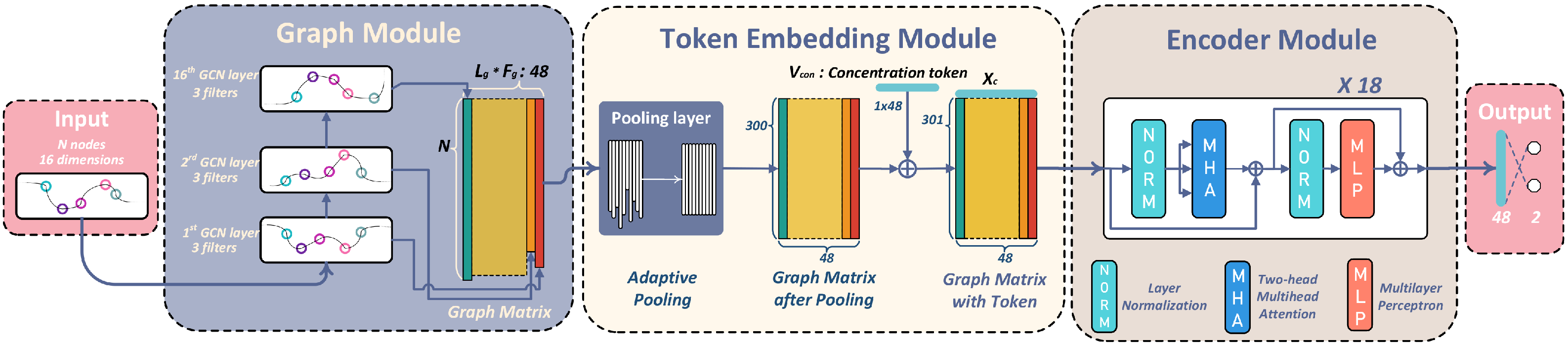}
   \caption{Architecture of the GraphANet model for concentration estimation.}
   \label{modelg}
\end{figure*}

As shown in Fig. \ref{modelg}, GraphANet consists of three main modules: the graph module, the token embedding module, and the encoder module. 
The graph module first processes sensor data graph $\mathcal{G}=(\textbf{X},\textbf{A} )$ using GCN to extract structural information, producing graph matrix, $\textbf{X}_{gs} \in \mathbb{R}^{N \times D}$, where $N$ denotes the number of nodes in the graph, and $D$ denotes the product of $L_{g}$ and $F_{g}$, $L_{g}$ is the number of GCN layers and $F_{g}$ is the number of GCN filters. Then, $N$ nodes are pooled into 300 nodes, forming the shape of each graph uniformly. A trainable concentration token $\textbf{v}_{con}$ is then stacked with graph matrix, where $\textbf{v}_{con} \in \mathbb{R}^{1 \times D}$, forming graph matrix with concentration token $\textbf{X}_{c}$ for the encoder. Then $\textbf{X}_{c}$ fed into the encoder module, which consists of $L$ identical blocks with two-head multihead attention and multilayer perceptron. Multihead attention is employed to capture the dependencies between elements across the sequence\cite{sa}, which has shown effective results in various gas recognition applications \cite{gsa1,gsa2,gsa3}, plays a key role in improving the performance of GraphANet for gas mixture concentration estimation. After extracting the class token $\textbf{v}_{con}$ into a fully connected layer's final head classifier, the concentration is determined.
GraphANet has parameters of $D$ = 48, $L_{g}$ = 16, $F_{g}$ = 3, $L$ = 18.   

We employ root mean square error (RMSE) as the loss function for the GraphANet model. 
\begin{equation}
RMSE = \sqrt{\frac{1}{2s} \sum_{i=1}^s \sum_{g=1}^2 (C_p^{ig} -C_t^{ig})^2}
\end{equation}

A lower RMSE value indicates a more accurate result, which also better fits the actual concentration regression curve, where $s$ denotes the number of samples in a batch, and $C_p^{ig}$ and $C_t^{ig}$ denote the prediction concentration and target concentration of the $g$-th gas component in the $i$-th sample, respectively.

\section{Experiment}

\subsection{Experimental setup}
We employed the same splitting method for both datasets. First, 16\% of the samples were set aside as the test set. The remaining samples were employed as the training and validation sets, employing a five-fold cross-validation method. The details of the dataset splitting are shown in Table \ref{data}. The model with the highest validation accuracy was selected for testing. To address the issue of sampling imbalance, stratified sampling was employed. The UCI test set contains 119 samples, while the custom dataset includes 96 samples.

In addition, since different gas has different concentration ranges, the concentrations are normalized by using the following formula:
\begin{equation}
y_i = \frac{y_o^{i}}{\max(y)}
\end{equation}
where $y_o^{i}$ and $\max(y)$ denote the $i$-class initial gas concentration and the maximum concentration of each gas in the entire dataset, respectively.

\subsection{Hyperparameter}
The hyperparameters for both models are displayed in Table \ref{H}. We set the training and validation batch sizes to 16 to accelerate the computation process. The test batch size is set as 4. Consistent hyperparameters were applied for training across both datasets.
\begin{table}[htb]
    \centering
    \caption{Hyperparameter.}
    \begin{tabular}{lc}
    \toprule
        Hyperparameter & GraphCapsNet/GraphANet  \\ \midrule
        Epochs & 200/1000 \\ 
        Training Batch Size & 16  \\ 
        Validation Batch Size & 16  \\ 
        Test Batch Size & 4 \\ 
        Learning Rate & 0.001  \\ 
        Optimizer & Adam \\ 
        Weight Decay & $10^{-6}$  \\ 
 \bottomrule
    \end{tabular}
  
    \label{H}
\end{table}

\subsection{Evaluation metrics}
\subsubsection{Classification Evaluation Metrics}
This study evaluates our multi-class classification model using accuracy, weighted precision, weighted recall, and weighted F1 score. Accuracy reflects the overall correctness of the model. Weighted precision and recall assess true positives relative to predicted and actual positives, respectively, averaging these metrics across classes and weighting them by the number of actual instances in each class. The weighted F1 score, calculated as the harmonic mean of precision and recall, provides a balanced evaluation by accounting for both false positives and false negatives, offering a comprehensive measure of the model's performance across all classes.\par

In addition, to comprehensively evaluate the classification model, we calculated the support-weighted accuracy by combining the accuracies of different datasets through sample weights. However, this approach does not fully reflect the model's performance, as the difficulty of identifying different datasets varies. The complexity arises from the number of gas types in the datasets and the presence of gas mixtures. Some datasets contain only pure gases, while others include mixtures, making them more challenging to classify. This complexity can be approximated by the increasing number of identifiable gas types, both pure and mixed. We proposed a comprehensive accuracy metric incorporating a gas variety weight into the support-weighted accuracy to address this. The formula for comprehensive accuracy is shown below:
\begin{equation}
W = \frac{\sum_{j=1}^{D} \left( f(C_j) \times \sqrt{S_j} \times A_j \right)}{\sum_{j=1}^{D} \left( f(C_j) \times \sqrt{S_j} \right)}
\end{equation} 
where $W$ is the comprehensive accuracy, $D$ is the number of datasets, $C_j$ is the number of gas types in the $j$-th dataset, $S_j$ is the support of the $j$-th dataset, and $A_j$ is the accuracy of the $j $-th dataset. The function $ f(C_j) $ is the degree of difficulty of the $j$-th dataset, it is defined as:

\begin{small}
\begin{equation}
f(C_j) = 
\begin{cases} 
C_j, & \text{if only pure gas} \\
\frac{\left( \sum_{k=2}^{C_j} N_k \times k \right) \times (2^{C_j} - C_j)}{C_j \left( 2^{C_j-1} - 1 \right)} + C_j , & \text{if partial mixtures} \\
2^{C_j}, & \text{if all mixtures} 
\end{cases}
\end{equation}
\end{small}
The parameteris employed for datasets with only pure gas, while $2^{C_j}$ applies to datasets includding all possible gas mixtures, and a more complex quadratic form is utilized for datasets containing partial mixtures, with values ranging between $C_j$ and $2^{C_j}$. Here, $k$ represents the number of gases in a particular mixture. For example, when considering ternary mixtures, $k = 3$. $N_k$ denotes the number of actual mixtures that contain exactly $k$ gases. The denominator of the formula represents the total number of possible gas components in all possible mixtures. At the same time, the numerator indicates the number of gas components present in the dataset's actual mixtures.

The parameter  $C_j$  is employed for datasets containing only pure gases, while $2^{C_j}$  applies to datasets including all possible gas mixtures. For datasets with partial mixtures, a more complex quadratic form is employed, with values ranging between $C_j$ and $2^{C_j}$. The variable $k$ denotes the number of gases in a specific mixture (e.g., for ternary mixtures, $k = 3$). $N_k$ denotes the number of actual mixtures containing exactly $k$ gases. The denominator in the formula reflects the total number of possible gas components across all potential mixtures, while the numerator corresponds to the number of gas components present in the actual mixtures within the dataset. For example, in the UCI dataset employed in this study, there are three gases: CO, CH\textsubscript{4}, and C\textsubscript{2}H\textsubscript{4}. The total number of possible gas components across all mixtures is calculated as $3 \times \left( 2^{3-1} - 1 \right) = 9$, resulting in a denominator of  9. However, since the dataset includes only the mixtures CO-C\textsubscript{2}H\textsubscript{4} and CH\textsubscript{4}-C\textsubscript{2}H\textsubscript{4}, the numerator is 4.

This distinction arises from the observation that for pure gases, the number of gas types increases linearly with the addition of each new gas. In contrast, for datasets containing all possible gas mixtures, including a new gas adds $2^n$ potential combinations, reflecting the combinatorial nature of gas mixtures. For datasets with only partial mixtures, the complexity of the mixtures is quantified by counting the occurrences of gas components across all possible combinations. Moreover, since the effect of sample size on model performance is not linear and the marginal benefits decrease as the sample size increases, the square root is employed to more accurately capture the actual impact of sample size. This approach prevents datasets with larger sample sizes from disproportionately influencing the overall accuracy and ensures a more balanced contribution from each dataset to the comprehensive accuracy.

\subsubsection{Gas concentration estimation evaluation metrics}
This study employs multiple metrics to comprehensively evaluate the model's performance, including the Coefficient of Determination ($\rm R^2$), Root Mean Squared Error (RMSE), Mean Absolute Error (MAE), Mean Absolute Percentage Error (MAPE), and Explained Variance Score (EV). The ($\rm R^2$) metric quantifies the proportion of variance in the dependent variable that can be explained by the independent variables, with values closer to 1 indicating a better model fit. RMSE measures the differences between predicted and actual values by calculating the square root of the mean squared errors; smaller values indicate higher prediction accuracy. MAE, which computes the average absolute differences between predicted and actual values, provides a straightforward measure of error size without amplifying larger errors. MAPE assesses prediction accuracy by expressing errors as a percentage of the actual values, making it particularly useful for comparing datasets with different scales, though it can be sensitive to small actual values. The EV score evaluates the model's ability to explain variability in the data, with higher scores reflecting superior explanatory power. By leveraging these metrics, the study offers an objective and multidimensional assessment of the regression model's predictive and explanatory performance.\par

\section{Results and Analysis}

\begin{table*}[htb]
    \caption{Gas mixture classification performance comparison of different models on UCI and custom datasets.}
    \label{tab:class performance}
    \centering
    \begin{tabular}{@{}lcccccccccc@{}}
        \toprule
        \multirow{2}{*}{Model} & \multicolumn{4}{c}{UCI dataset} & \multicolumn{4}{c}{Custom dataset} & \multirow{2}{*}{\shortstack{Support \\weighted accuracy}} & \multirow{2}{*}{\shortstack{Comprehensive \\accuracy}} \\
        \cmidrule(lr){2-5} \cmidrule(lr){6-9}
        & Accuracy & Precision & Recall & F1score & Acc & Precision & Recall & F1score &  &  \\
        \midrule
        GraphCapsNet (ours)   & \textbf{0.983} & \textbf{0.984} & \textbf{0.983} & \textbf{0.983} & \textbf{0.990} & \textbf{0.990} & \textbf{0.990} & \textbf{0.990} & \textbf{0.986} & \textbf{0.986} \\\midrule
        PSCFormer\cite{pscformer} & \ul{0.958} & \ul{0.960} & \ul{0.958} & \ul{0.957} & \ul{0.979} & \ul{0.979} & \ul{0.979} & \ul{0.979} & \ul{0.967} & \ul{0.967} \\\midrule
        TeTCN\cite{TeTCN}     & 0.832 & 0.832 & 0.832 & 0.831 & 0.969 & 0.971 & 0.969 & 0.969 & 0.893 & 0.888 \\\midrule
        SVM       & 0.931 & 0.934 & 0.931 & 0.931 & 0.969 & 0.970 & 0.969 & 0.969 & 0.948 & 0.946 \\\midrule
        KNN       & 0.931 & 0.934 & 0.931 & 0.931 & 0.958 & 0.964 & 0.958 & 0.958 & 0.943 & 0.942 \\\midrule
        RF        & 0.922 & 0.926 & 0.922 & 0.923 & \ul{0.979} & \ul{0.979} & \ul{0.979} & \ul{0.979} & 0.948 & 0.945 \\
        \bottomrule
    \end{tabular}
\end{table*}

\begin{table*}[!ht]
\caption{Performance metrics of various concentration estimation models on different groups.}
\label{rrt}
\centering
\begin{adjustbox}{width=1\textwidth,center}
\begin{tabularx}{\textwidth}{@{}l*{11}{>{\centering\arraybackslash}X}@{}}
\toprule

\multicolumn{12}{c}{\textbf{UCI Group A}} \\ \midrule
\multirow{2}{*}{Algorithm} & \multicolumn{3}{c}{$\rm R^2$ (CO)} & \multicolumn{3}{c}{$\rm R^2$ (C\textsubscript{2}H\textsubscript{4})} & \multirow{2}{*}{$\rm R^2$} & \multirow{2}{*}{MAE} & \multirow{2}{*}{RMSE} & \multirow{2}{*}{MAPE (\%)} & \multirow{2}{*}{EV}   \\ \cmidrule(lr){2-4}\cmidrule(lr){5-7}
& Pure & Mixed & Total & Pure & Mixed & Total & & & \\ \midrule
GraphANet(ours)   & \textbf{0.986} & \textbf{0.913}  & \textbf{0.975} & 0.960 & \textbf{0.956}  & \textbf{0.959} & \textbf{0.965} & \textbf{0.023} & \textbf{0.030} & \textbf{4.24}  & \textbf{0.965}  \\ \midrule
2TCN\cite{2TCN}   & 0.288 & 0.316  & 0.326 & \textbf{0.988} & 0.525  & 0.901 & 0.782 & \ul{0.036} & \ul{0.062} & 7.37  & 0.795  \\ \midrule
LSTMA\cite{LSTMA}  & 0.623 & \ul{0.326}  & 0.633 & 0.617 & 0.738  & 0.651 & 0.665 & 0.057 & 0.085 & 10.72 & 0.666  \\ \midrule
ViT\cite{vit}    & 0.660 & -0.318 & 0.518 & 0.513 & -0.104 & 0.349 & 0.407 & 0.079 & 0.124 & 14.48 & 0.420  \\ \midrule
SVR    & 0.603 & 0.059  & 0.617 & 0.829 & 0.632  & 0.810 & 0.716 & 0.062 & 0.079 & 11.80 & 0.717  \\ \midrule
KNN    & 0.479 & -0.087 & 0.365 & 0.494 & 0.637  & 0.568 & 0.520 & 0.053 & 0.109 & 9.35  & 0.546  \\ \midrule
RF     & \ul{0.933} & 0.220  & \ul{0.761} & \ul{0.970} & \ul{0.744}  & \ul{0.907} & \ul{0.839} & 0.040 & 0.073 & \ul{6.69}  & \ul{0.847}  \\ 
\bottomrule
\toprule
\multicolumn{12}{c}{\textbf{UCI Group B}} \\ \midrule
\multirow{2}{*}{Algorithm} & \multicolumn{3}{c}{$\rm R^2$ (CH\textsubscript{4})} & \multicolumn{3}{c}{$\rm R^2$ (C\textsubscript{2}H\textsubscript{4})} & \multirow{2}{*}{$\rm R^2$} & \multirow{2}{*}{MAE} & \multirow{2}{*}{RMSE} & \multirow{2}{*}{MAPE (\%)} & \multirow{2}{*}{EV} \\ \cmidrule(lr){2-4}\cmidrule(lr){5-7}
& Pure & Mixed & Total & Pure & Mixed & Total & & & \\ \midrule
GraphANet(ours)  & \textbf{0.997} & \ul{0.964} & \textbf{0.992} & 0.988 & \textbf{0.980}  & \textbf{0.989} & \textbf{0.991} & \textbf{0.014} & \textbf{0.019} & \textbf{3.11}  & \textbf{0.991} \\ \midrule
2TCN\cite{2TCN}   & \ul{0.987} & 0.493 & \ul{0.981} & 0.984 & \ul{0.927}  & 0.936 & 0.952 & 0.023 & 0.044 & 5.97  & 0.956 \\ \midrule
LSTMA\cite{LSTMA}  & 0.823 & 0.678 & 0.788 & 0.919 & 0.116  & 0.898 & 0.870 & 0.049 & 0.061 & 10.90 & 0.870 \\ \midrule
ViT\cite{vit}    & 0.893 & 0.352 & 0.794 & 0.796 & 0.351  & 0.843 & 0.833 & 0.053 & 0.074 & 12.54 & 0.840 \\ \midrule
SVR    & 0.813 & 0.453 & 0.803 & 0.391 & -0.392 & 0.546 & 0.689 & 0.070 & 0.113 & 16.19 & 0.694 \\ \midrule
KNN    & 0.799 & \textbf{0.983} & 0.873 & \ul{0.992} & 0.842  & 0.978 & 0.919 & \ul{0.022} & 0.051 & \ul{4.66}  & 0.922 \\ \midrule
RF     & 0.950 & 0.870 & 0.948 & \textbf{0.993} & 0.882  & \ul{0.983} & \ul{0.962} & \ul{0.022} & \ul{0.037} & 5.62  & \ul{0.963} \\ 
\bottomrule
\toprule
\multicolumn{12}{c}{\textbf{Custom Dataset}} \\ \midrule
\multirow{2}{*}{Algorithm} & \multicolumn{3}{c}{$\rm R^2$ (H\textsubscript{2})} & \multicolumn{3}{c}{$\rm R^2$ (C\textsubscript{2}H\textsubscript{4})} & \multirow{2}{*}{$\rm R^2$} & \multirow{2}{*}{MAE} & \multirow{2}{*}{RMSE} & \multirow{2}{*}{MAPE (\%)} & \multirow{2}{*}{EV} \\ \cmidrule(lr){2-4}\cmidrule(lr){5-7}
& Pure & Mixed & Total & Pure & Mixed & Total & & & \\ \midrule
GraphANet(ours)  & \ul{0.990} & \ul{0.989} & \ul{0.990} & 0.971 & \textbf{0.907} & \textbf{0.933} & \textbf{0.960} & \textbf{0.037} & \textbf{0.054} & \ul{11.53} & \textbf{0.961} \\ \midrule
2TCN\cite{2TCN}  & 0.986 & 0.972 & 0.978 & 0.929 & \ul{0.888} & \ul{0.904} & 0.940 & 0.047 & 0.066 & 14.79 & 0.942 \\ \midrule
LSTMA\cite{LSTMA} & \textbf{0.997} & 0.974 & 0.984 & \textbf{0.996} & 0.756 & 0.854 & 0.918 & 0.043 & 0.077 & 12.55 & 0.918 \\ \midrule
ViT\cite{vit}   & 0.988 & \textbf{0.992} & \textbf{0.991} & \ul{0.984} & 0.846 & 0.901 & \ul{0.945} & \ul{0.039} & \ul{0.064} & \textbf{11.02} & \ul{0.945} \\ \midrule
SVR   & 0.481 & 0.850 & 0.724 & 0.366 & 0.459 & 0.423 & 0.571 & 0.131 & 0.175 & 35.28 & 0.612 \\ \midrule
KNN   & 0.699 & 0.937 & 0.854 & 0.830 & 0.687 & 0.746 & 0.800 & 0.084 & 0.120 & 24.48 & 0.804 \\ \midrule
RF    & 0.794 & 0.932 & 0.885 & 0.696 & 0.733 & 0.719 & 0.801 & 0.079 & 0.119 & 20.79 & 0.808 \\ 
\bottomrule
\end{tabularx}
\end{adjustbox}
\end{table*}

To validate the performance of our models, we conducted a comprehensive comparative evaluation involving GraphCapsNet and GraphANet. For the gas mixture identification task, we compared our models with recently published approaches, including PSCFormer \cite{pscformer} and TeTCN \cite{TeTCN}. For the concentration identification task, we benchmarked against two-channel TCN (2TCN) \cite{2TCN} and LSTMA \cite{LSTMA}. Additionally, ablation experiments were conducted using the vision transformer (ViT) model \cite{vit} to assess the contributions of individual components. Beyond deep learning models, we also incorporated traditional machine learning approaches, including Support Vector Machines (SVM), Random Forests (RF), and K-Nearest Neighbors (KNN), to evaluate both tasks, providing a thorough performance comparison.
\subsection{Overview of classification results}
\begin{figure*}[!h]
\centering
	\subfigure[]
	{\begin{minipage}[c]{0.24\textwidth}
      \centering
      \includegraphics[width=\textwidth]{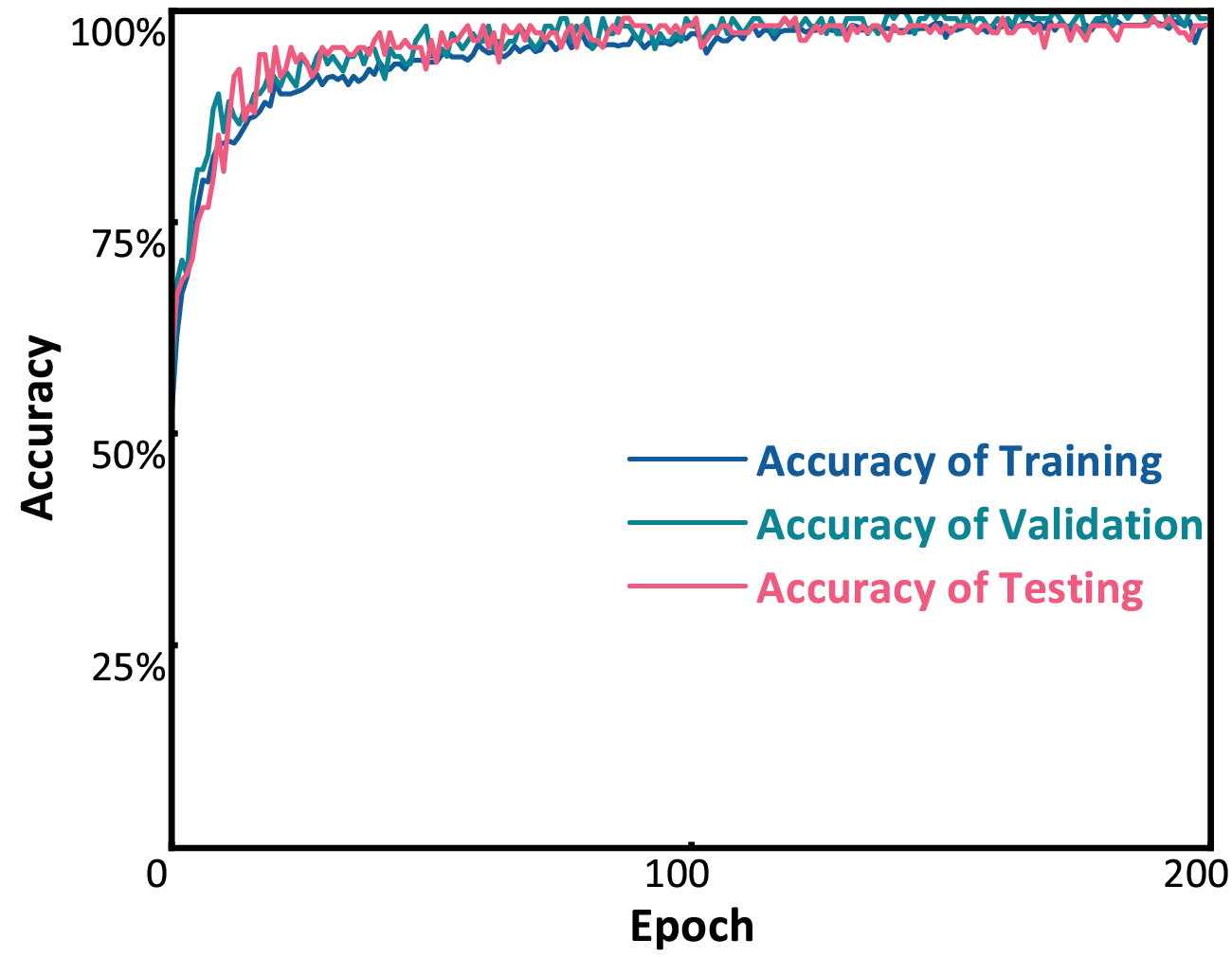}
      \end{minipage}}
     	\subfigure[]
	{\begin{minipage}[c]{0.24\textwidth}
	 \centering
      \includegraphics[width=\textwidth]{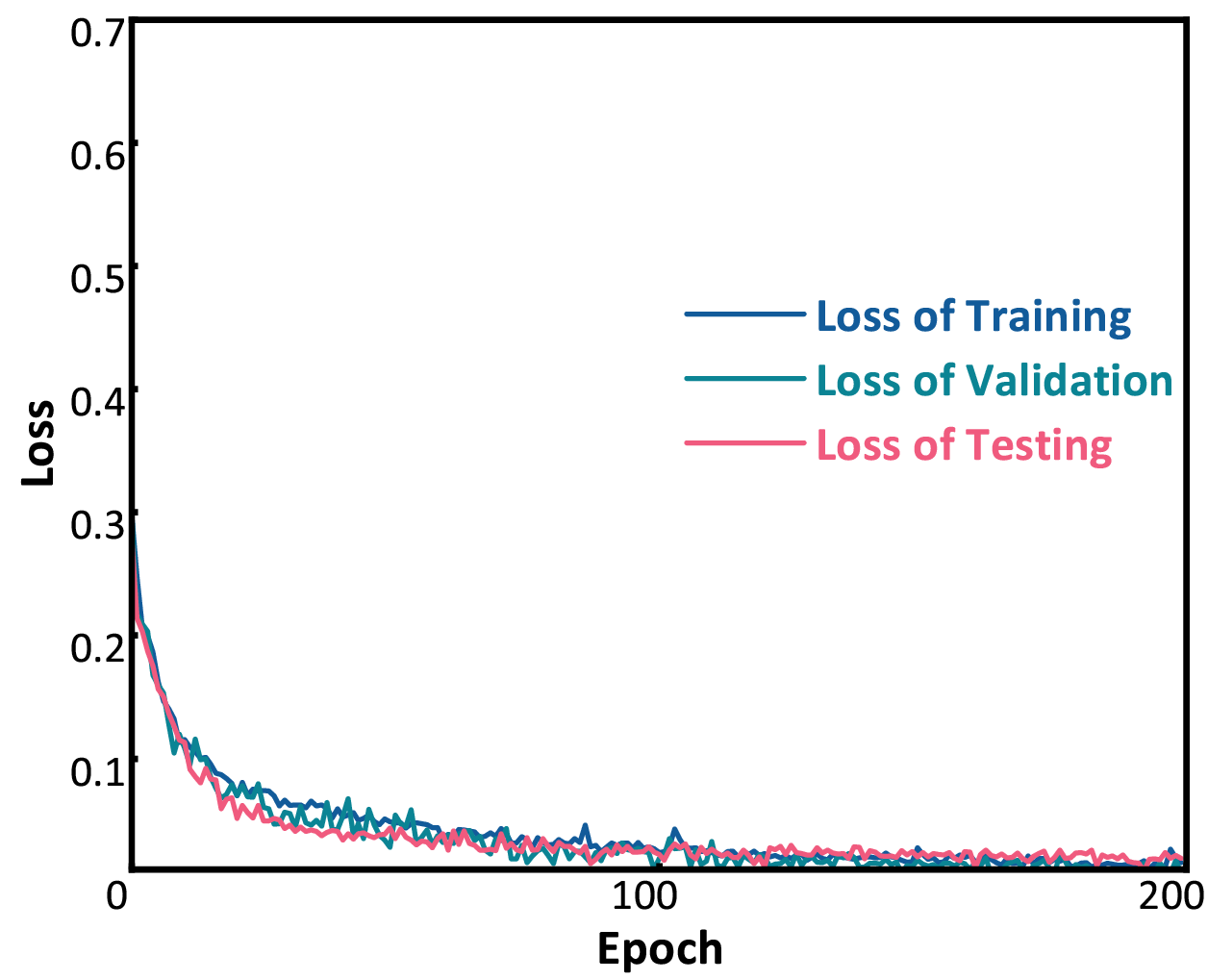}
      \end{minipage}}
     	\subfigure[]
	{\begin{minipage}[c]{0.23\textwidth}
	 \centering
      \includegraphics[width=\textwidth]{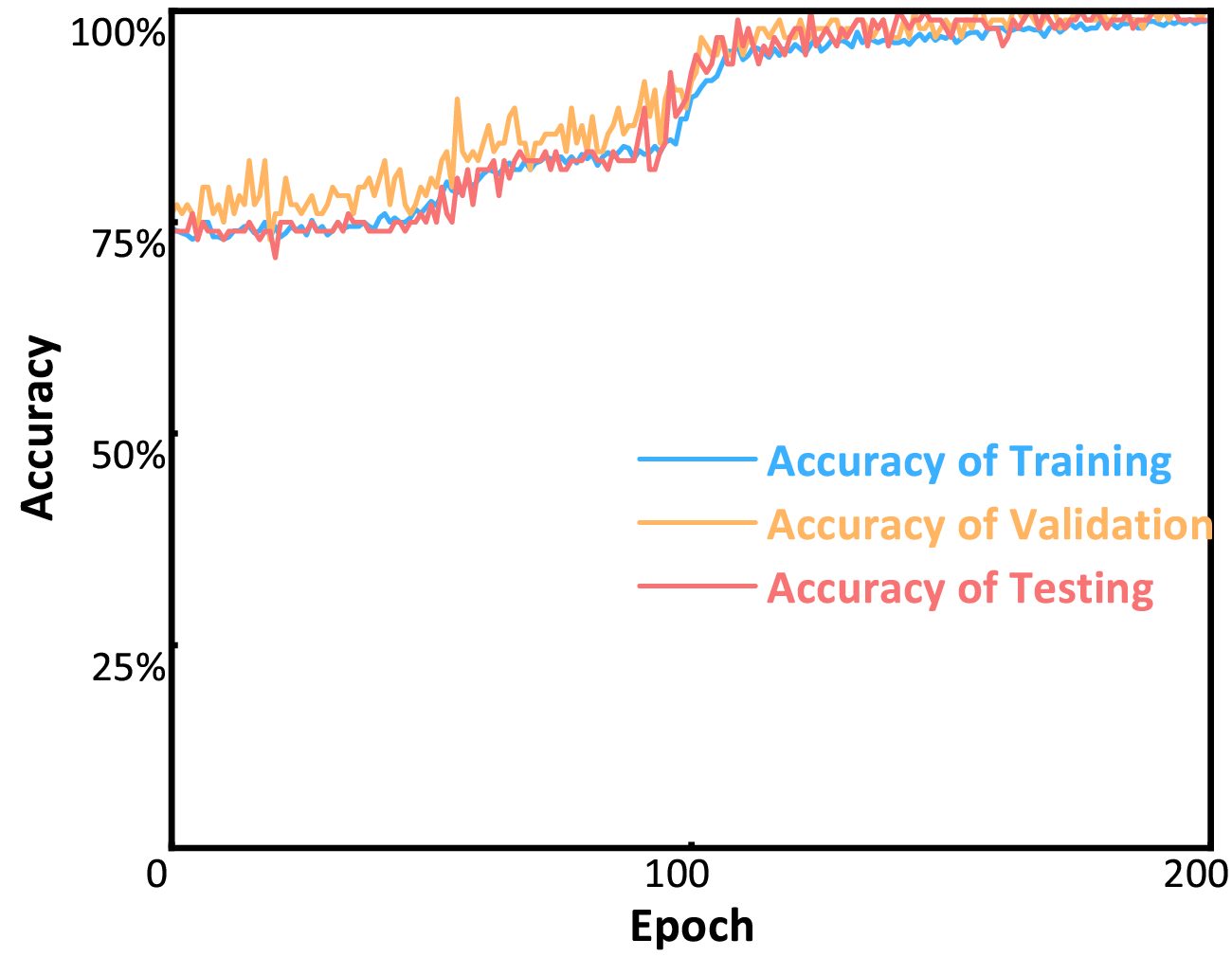}
      \end{minipage}}
	\subfigure[]
	{\begin{minipage}[c]{0.23\textwidth}
	 \centering
      \includegraphics[width=\textwidth]{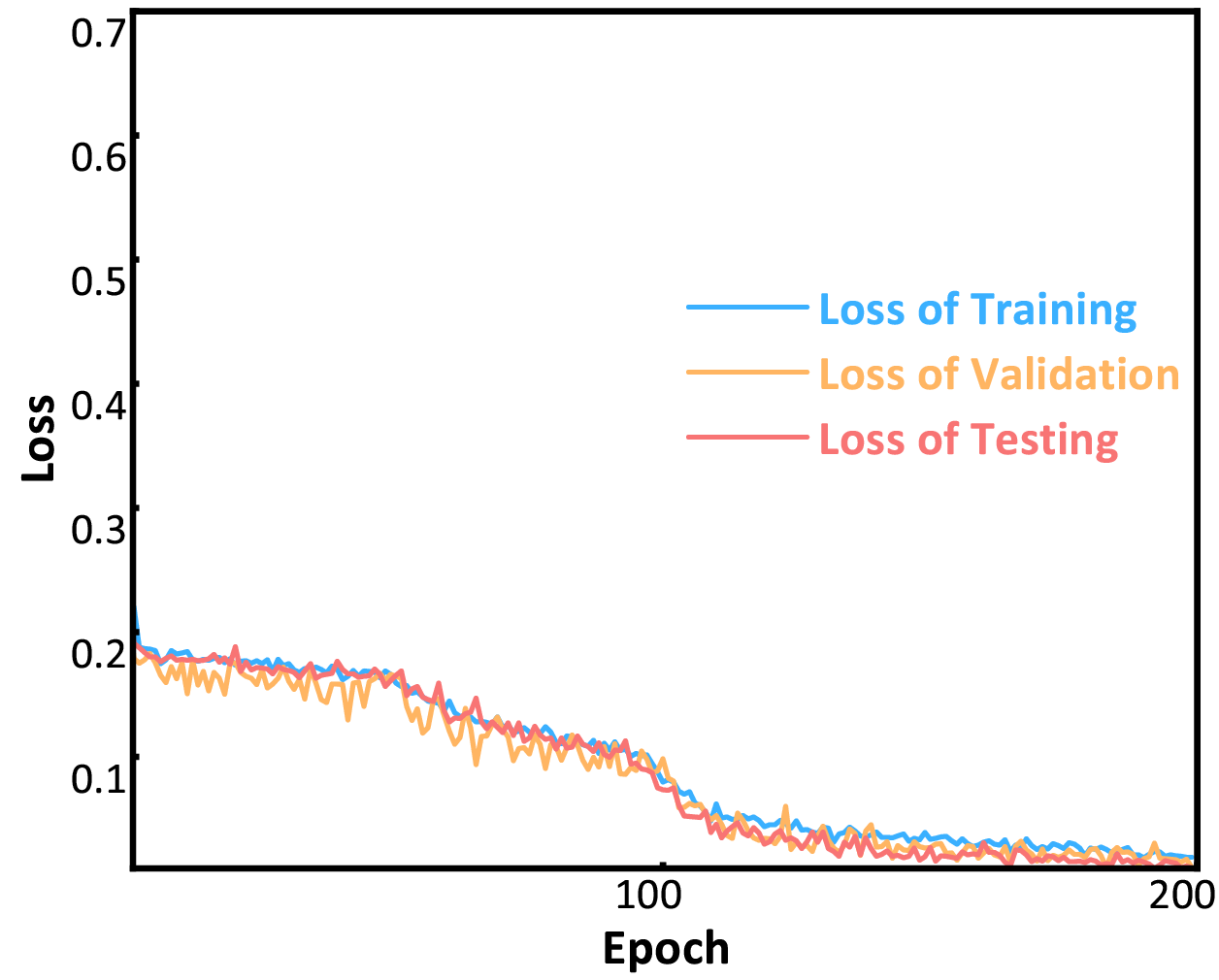}
      \end{minipage}}
\caption{The results of gas identification and concentration estimation. (a) Training, validation, and testing accuracy during 200 epochs on the UCI dataset. (b) The loss of training, validation, and testing during 200 epochs on the UCI dataset. (c) The accuracy of training, validation, and testing during 200 epochs on the custom dataset. (d) The loss of training, validation, and testing during 200 epochs on custom dataset.}
\label{result}
\end{figure*}

\begin{figure*}[!ht]
\centering
	\subfigure[GraphCapsNet]
	{\begin{minipage}[c]{0.3\textwidth}
      \centering
      \includegraphics[width=\textwidth]{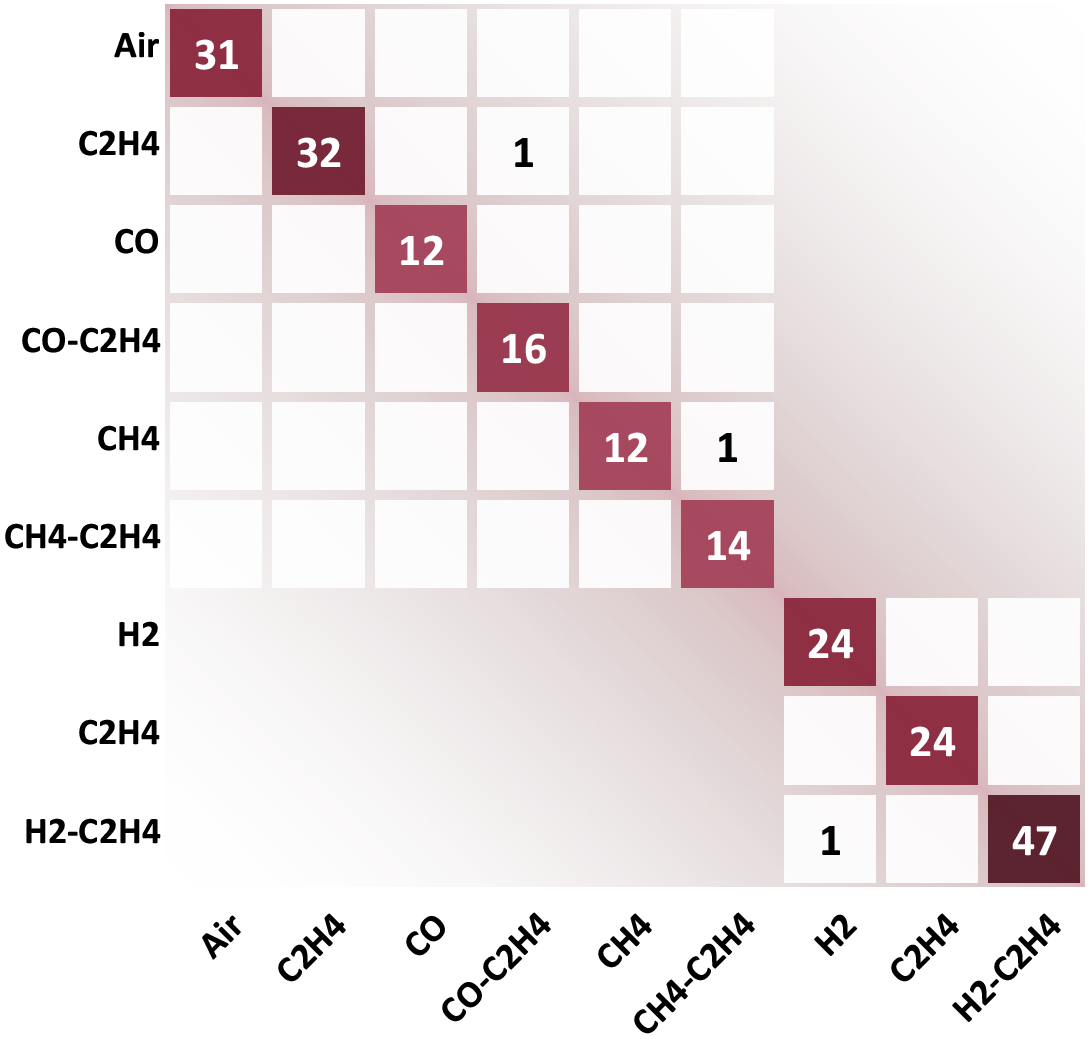}
      \end{minipage}}
	\subfigure[PSCFormer]
	{\begin{minipage}[c]{0.3\textwidth}
	 \centering
      \includegraphics[width=\textwidth]{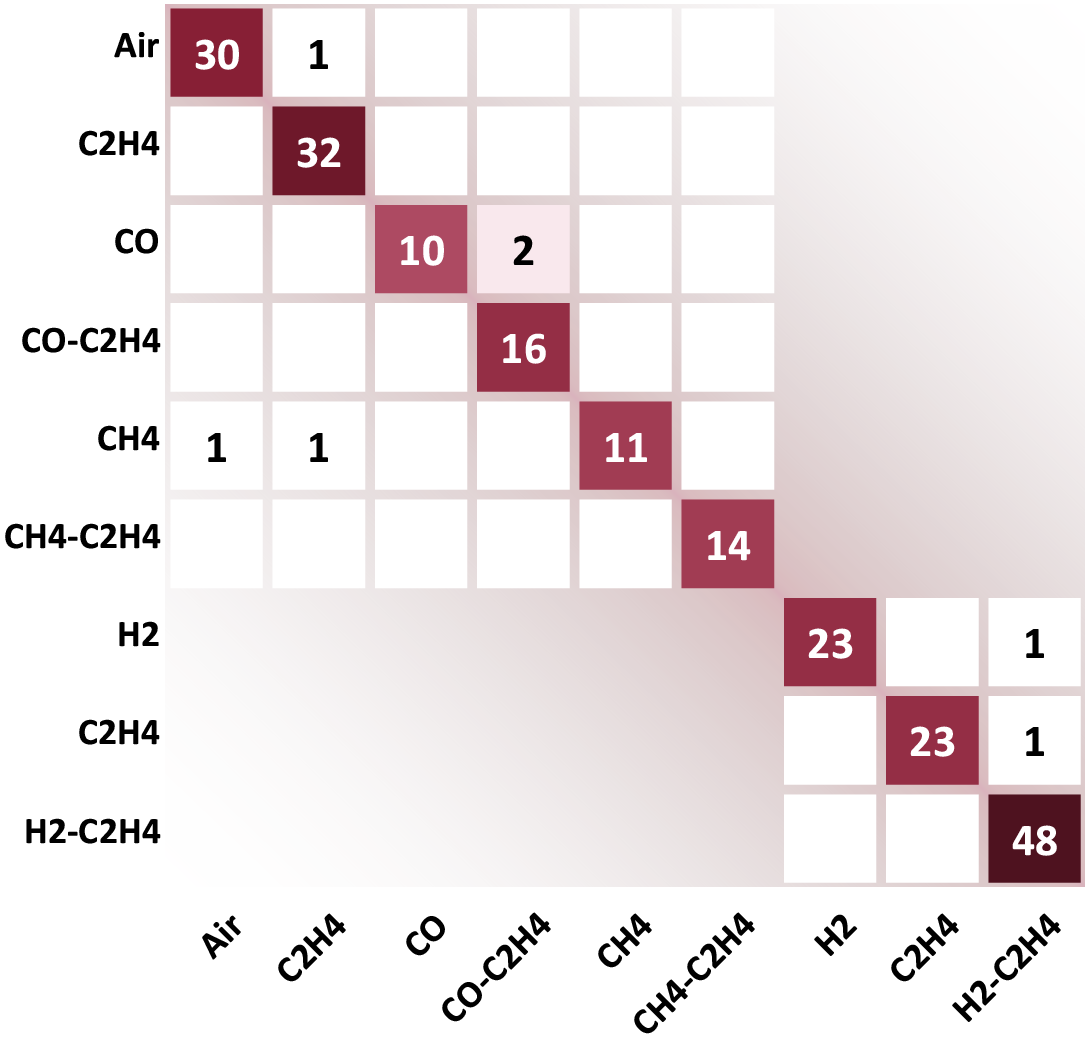}
      \end{minipage}}
	\subfigure[TeTCN]
      {\begin{minipage}[c]{0.3\textwidth}
	 \centering
      \includegraphics[width=\textwidth]{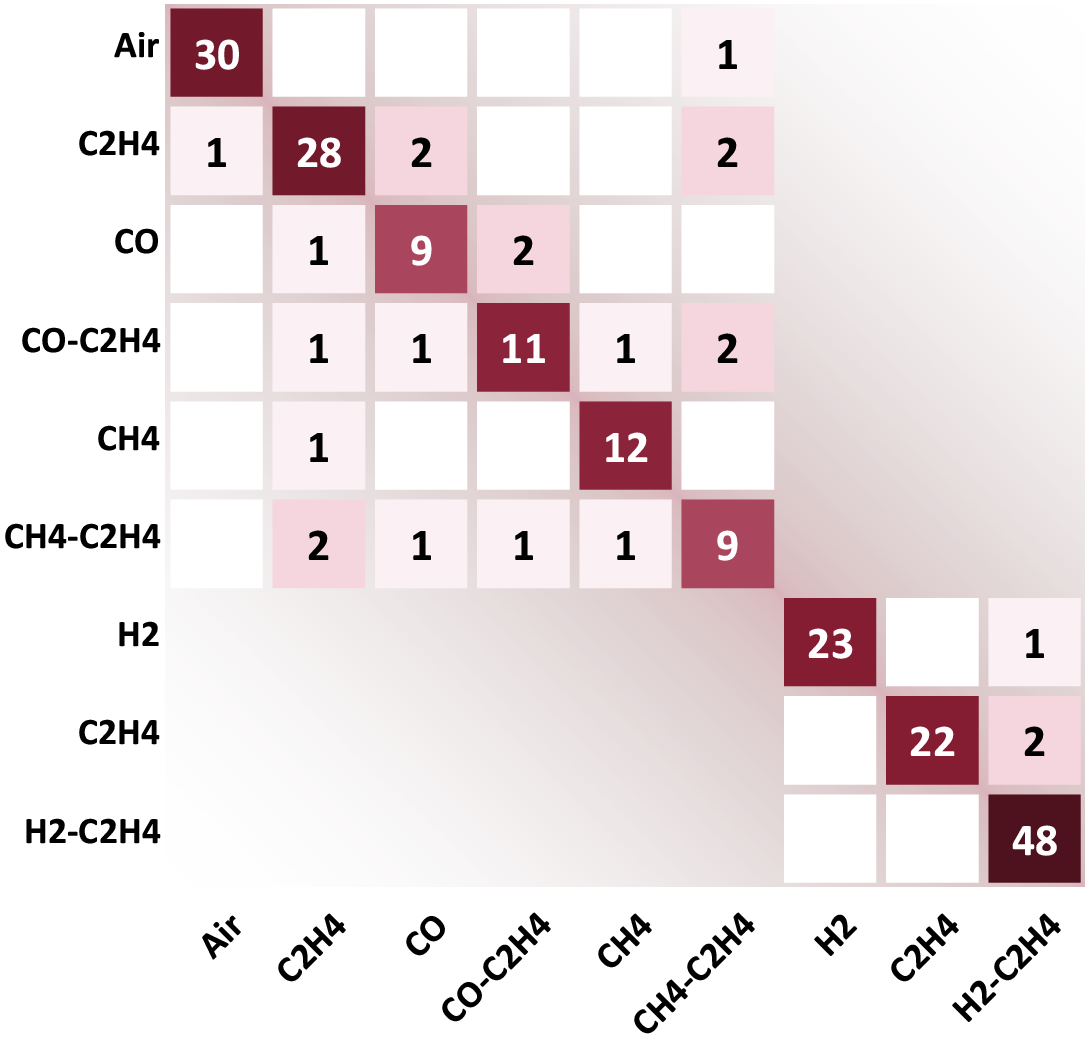}
      \end{minipage}}
\caption{The confusion matrix of GraphCapsNet, PSCFormer, and TeTCN on two datasets. The 6$\times$6 matrix in the upper left corner of each confusion matrix represents the results from experiments on the UCI dataset, while the 3$\times$3 matrix in the lower right corner represents the results from experiments on the custom dataset.}
\label{figcm}
\end{figure*}

\begin{figure*}[!ht]
	\subfigure[GraphANet/CO-C\textsubscript{2}H\textsubscript{4}]
	{\begin{minipage}[c]{0.24\textwidth}
      \centering
      \includegraphics[width=\textwidth]{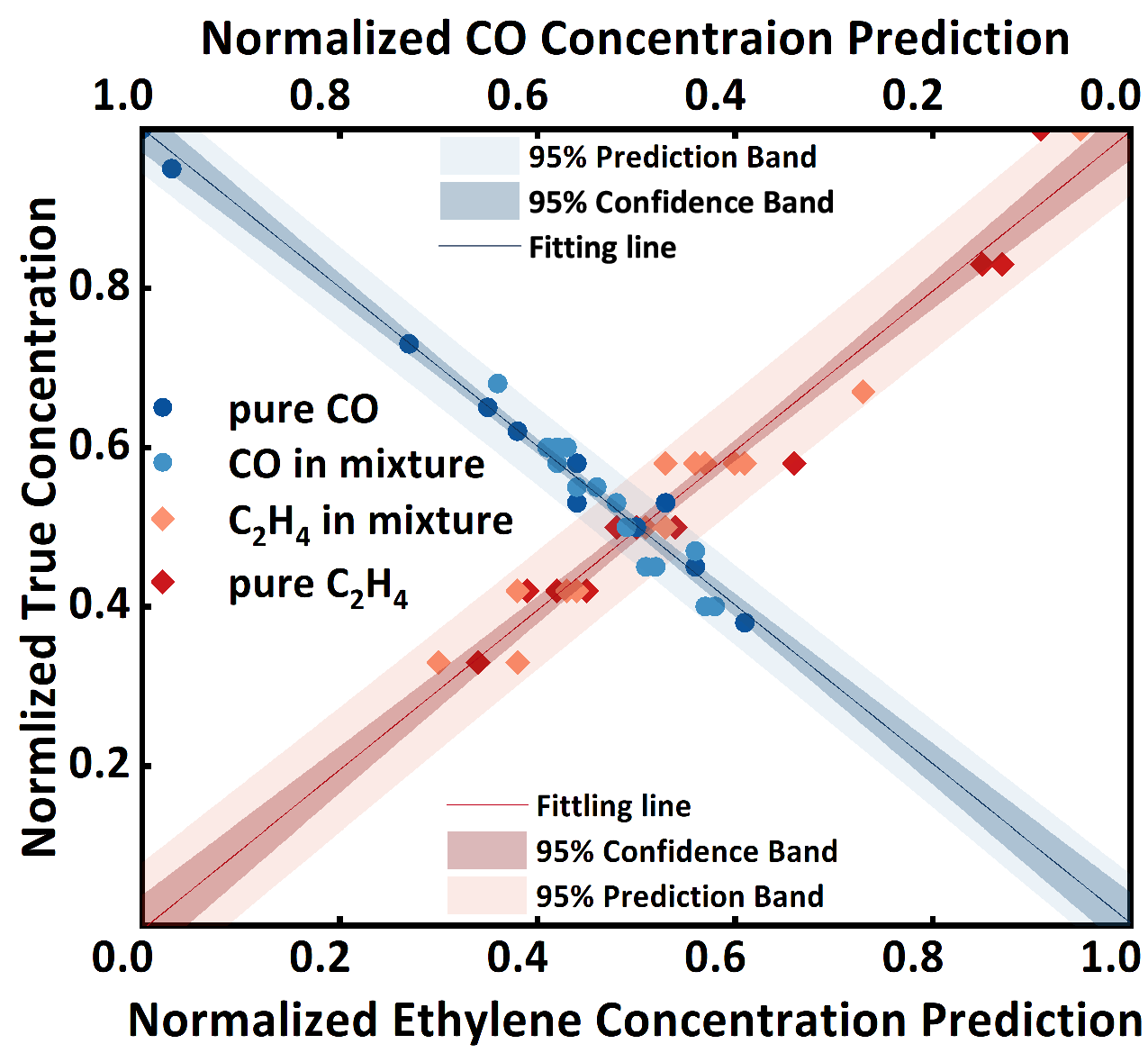}
      \end{minipage}}
        \subfigure[LSTMA/CO-C\textsubscript{2}H\textsubscript{4}]
	{\begin{minipage}[c]{0.24\textwidth}
      \centering
      \includegraphics[width=\textwidth]{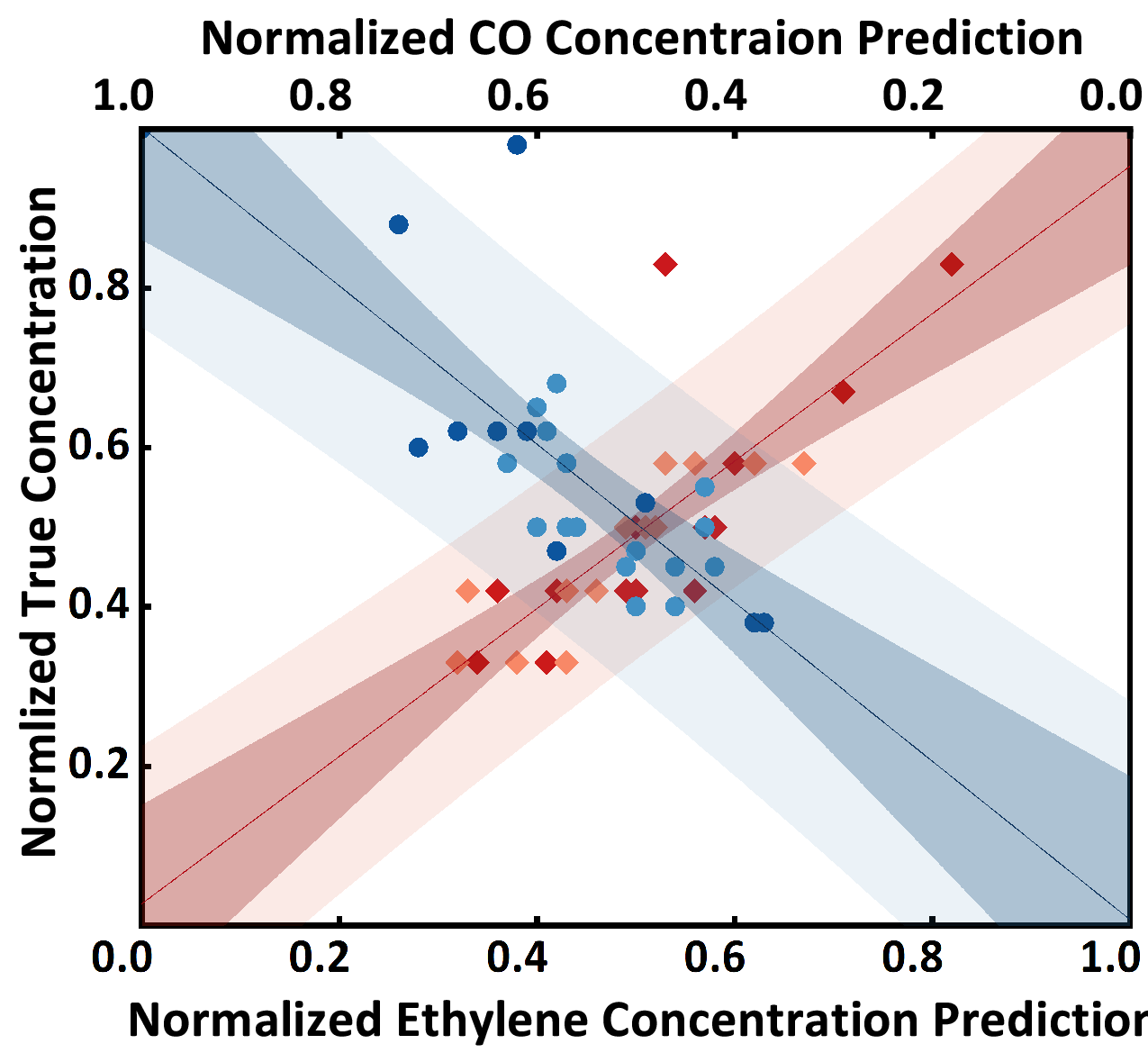}
      \end{minipage}}
        \subfigure[2TCN/CO-C\textsubscript{2}H\textsubscript{4}]
	{\begin{minipage}[c]{0.24\textwidth}
      \centering
      \includegraphics[width=\textwidth]{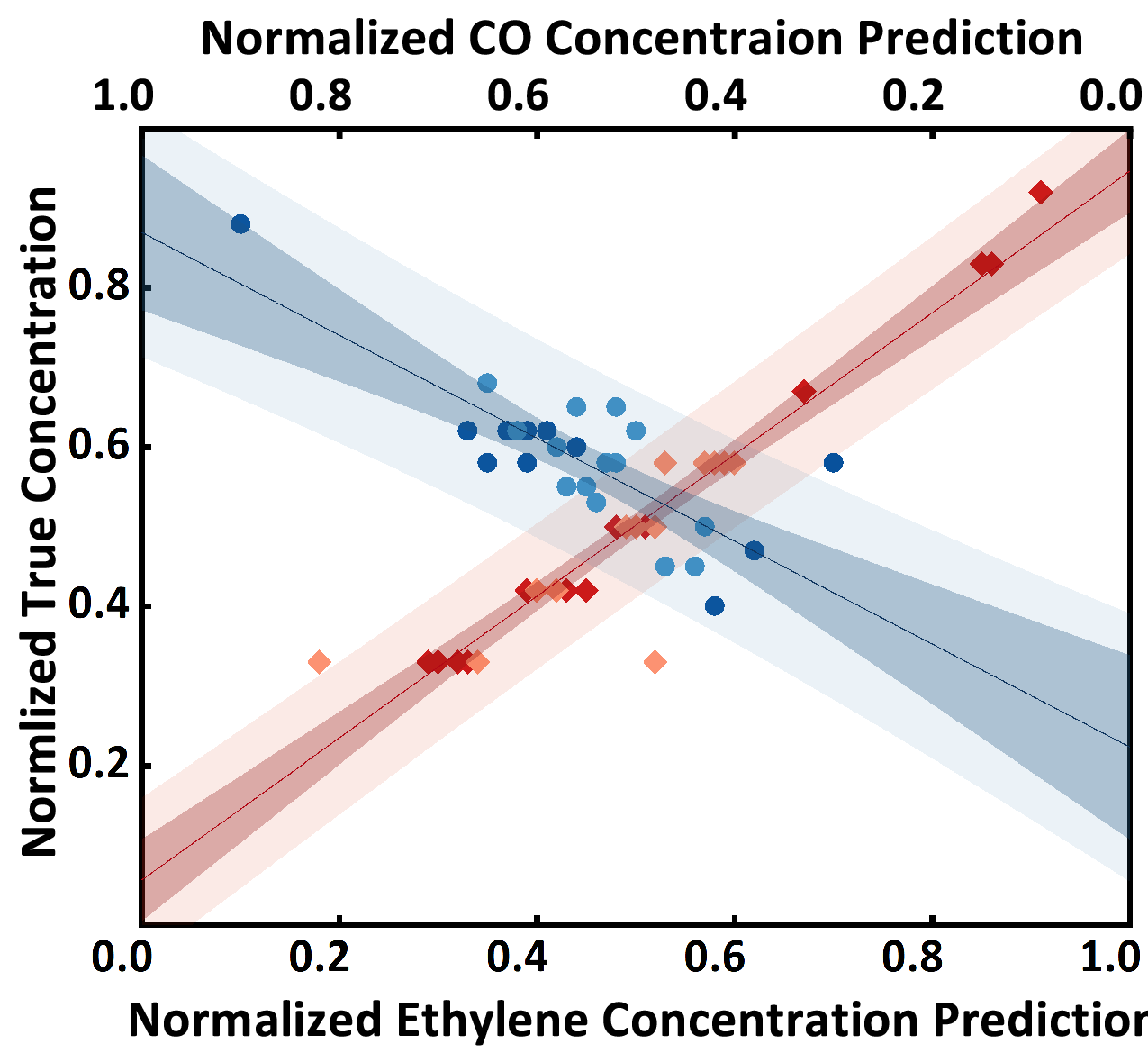}
      \end{minipage}}
      \subfigure[ViT/CO-C\textsubscript{2}H\textsubscript{4}]
	{\begin{minipage}[c]{0.24\textwidth}
      \centering
      \includegraphics[width=\textwidth]{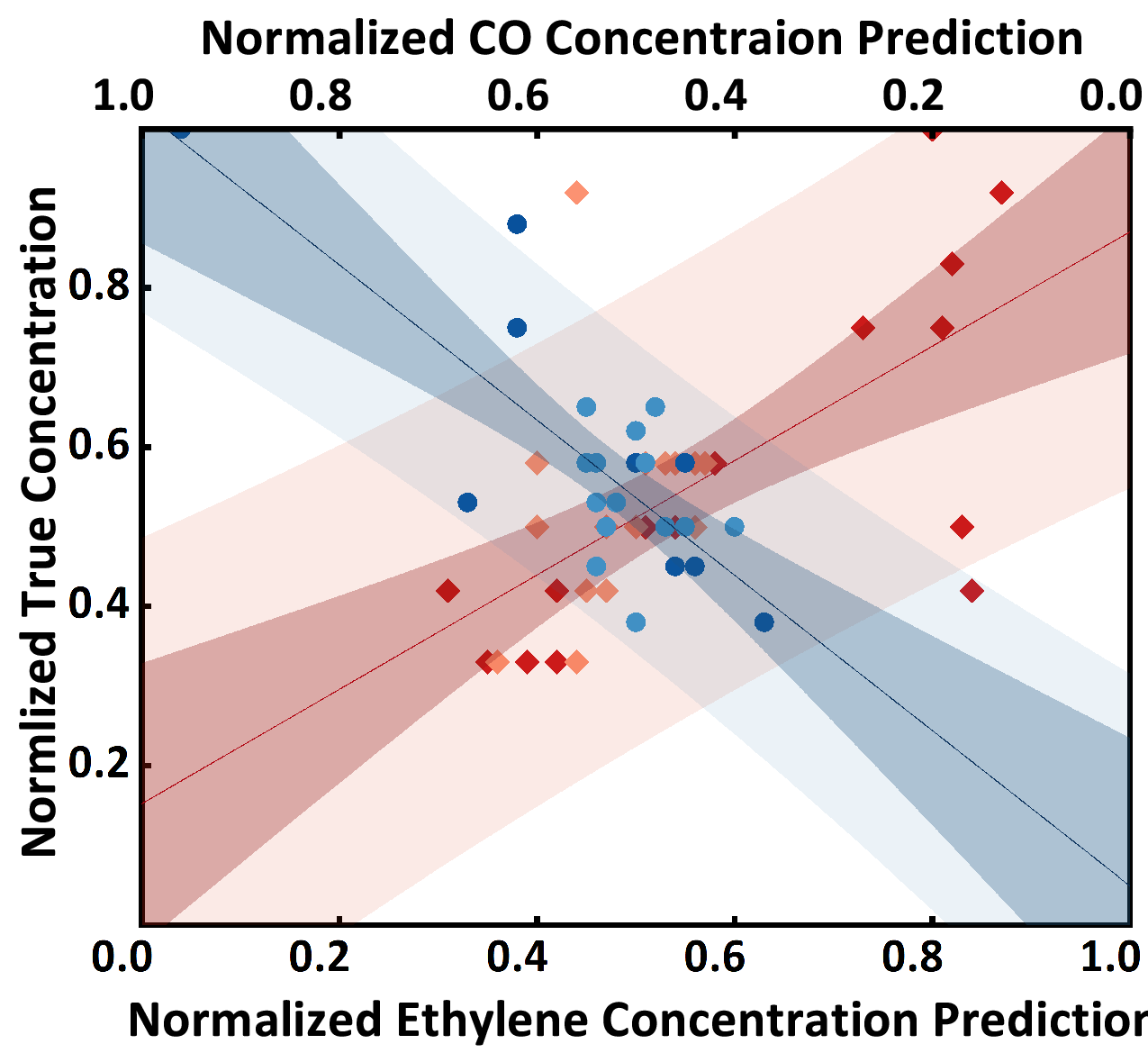}
      \end{minipage}}   
	\subfigure[GraphANet/CH\textsubscript{4}-C\textsubscript{2}H\textsubscript{4}]
	{\begin{minipage}[c]{0.24\textwidth}
	 \centering
      \includegraphics[width=\textwidth]{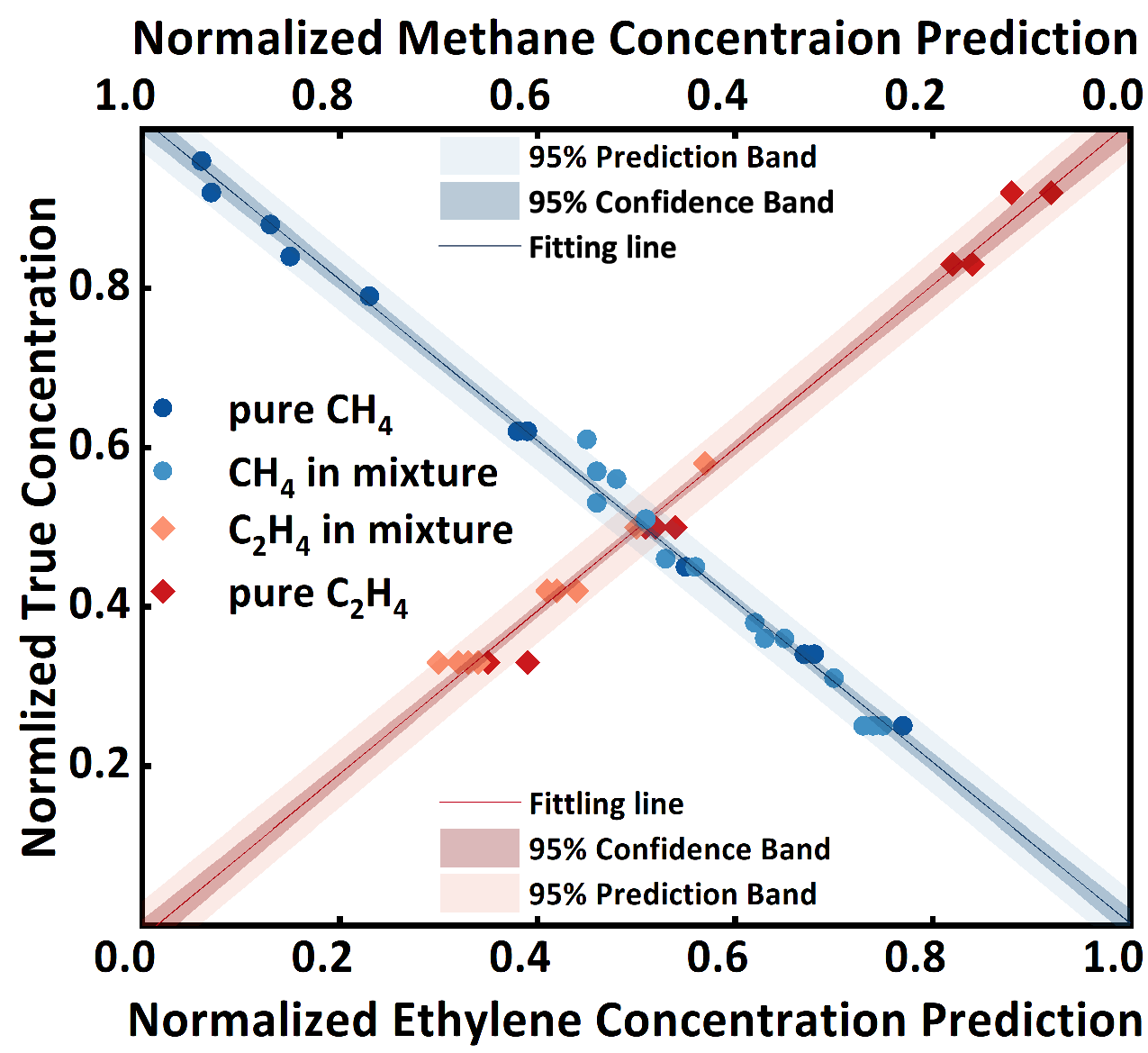}
      \end{minipage}}
      \subfigure[LSTMA/CH\textsubscript{4}-C\textsubscript{2}H\textsubscript{4}]
	{\begin{minipage}[c]{0.24\textwidth}
	 \centering
      \includegraphics[width=\textwidth]{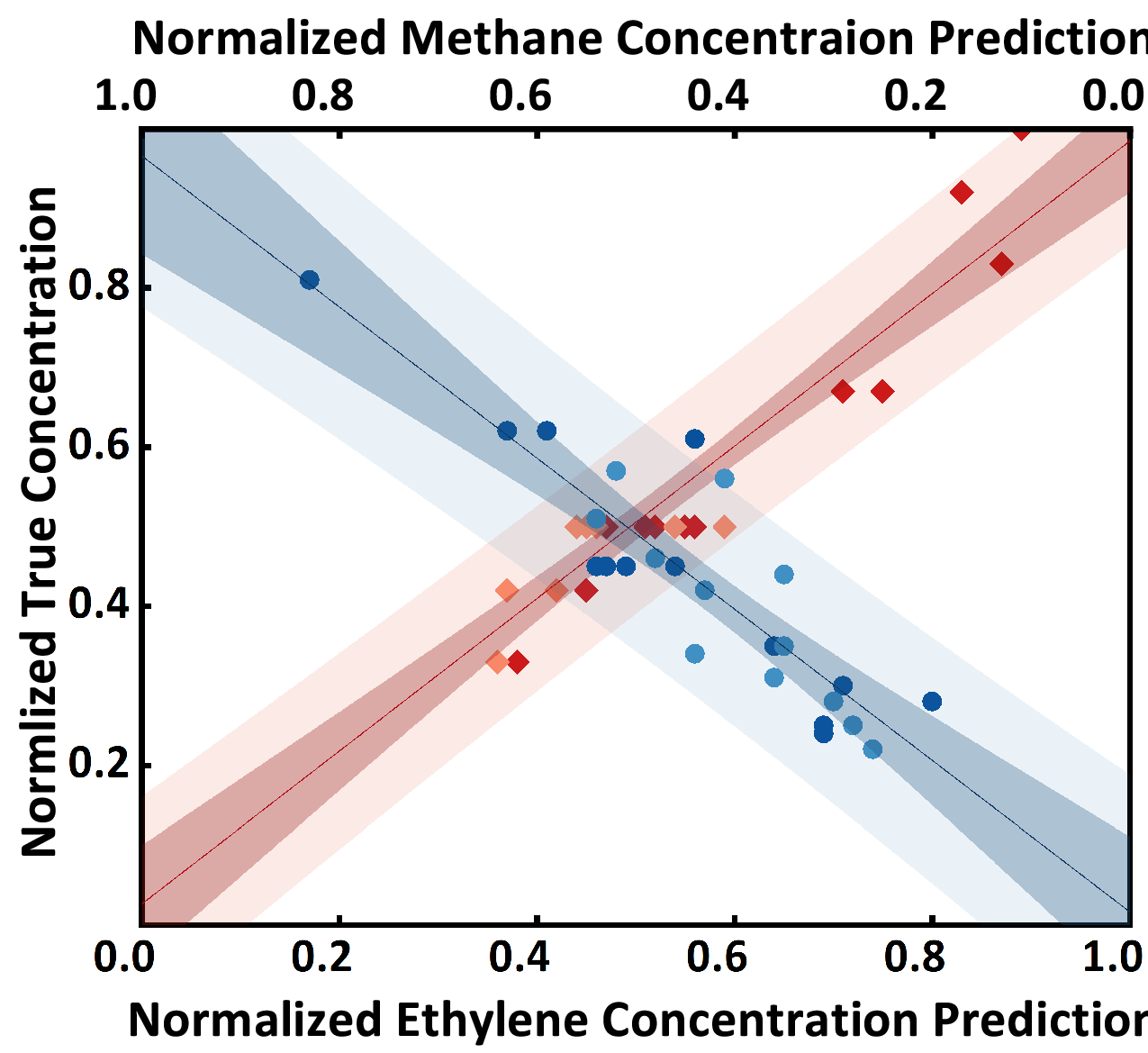}
      \end{minipage}}
      \subfigure[2TCN/CH\textsubscript{4}-C\textsubscript{2}H\textsubscript{4}]
	{\begin{minipage}[c]{0.24\textwidth}
	 \centering
      \includegraphics[width=\textwidth]{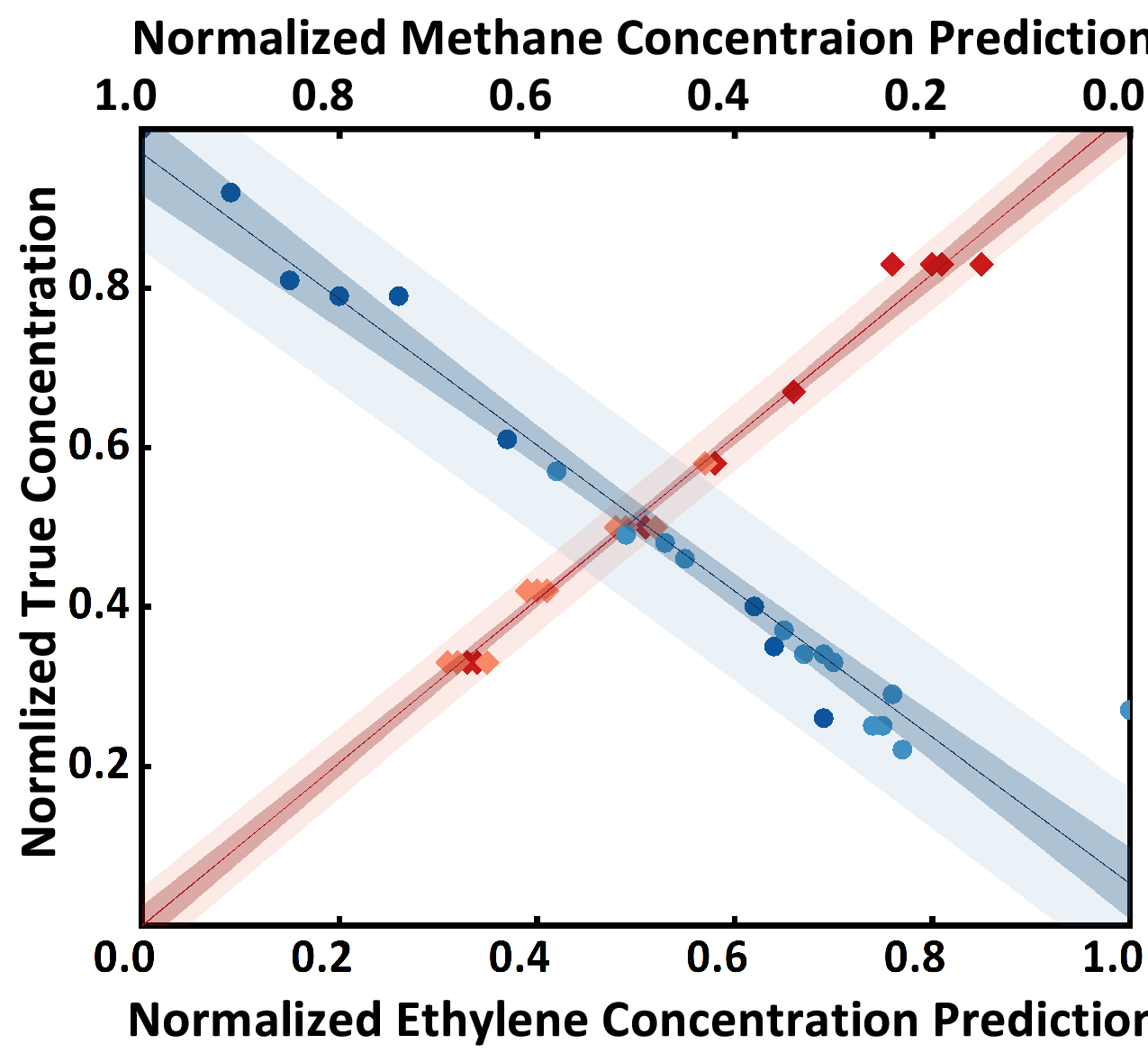}
      \end{minipage}}
      \subfigure[ViT/CH\textsubscript{4}-C\textsubscript{2}H\textsubscript{4}]
	{\begin{minipage}[c]{0.24\textwidth}
	 \centering
      \includegraphics[width=\textwidth]{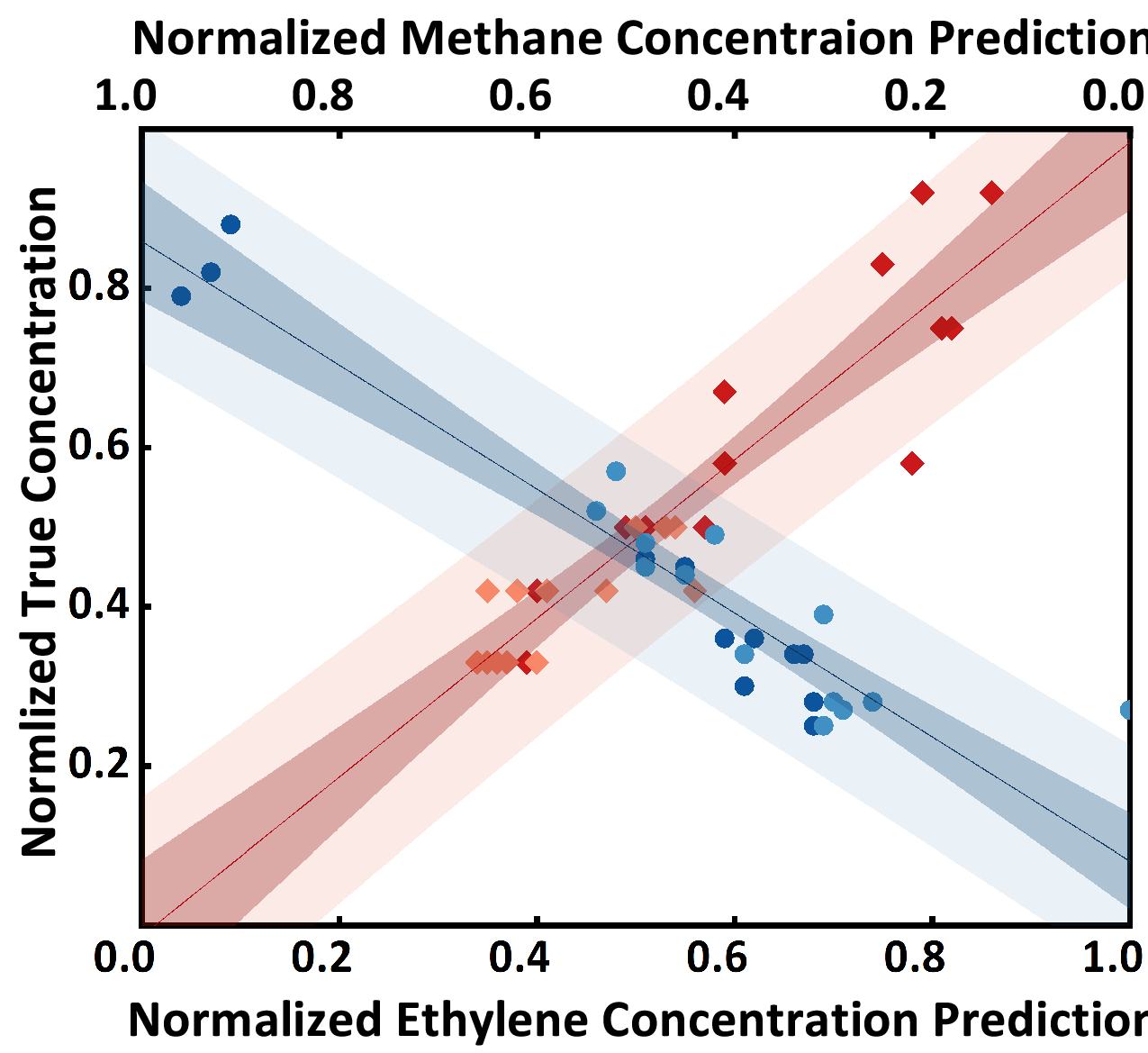}
      \end{minipage}}
      \subfigure[GraphANet/H\textsubscript{2}-C\textsubscript{2}H\textsubscript{4}]
      {\begin{minipage}[c]{0.24\textwidth}
	 \centering
      \includegraphics[width=\textwidth]{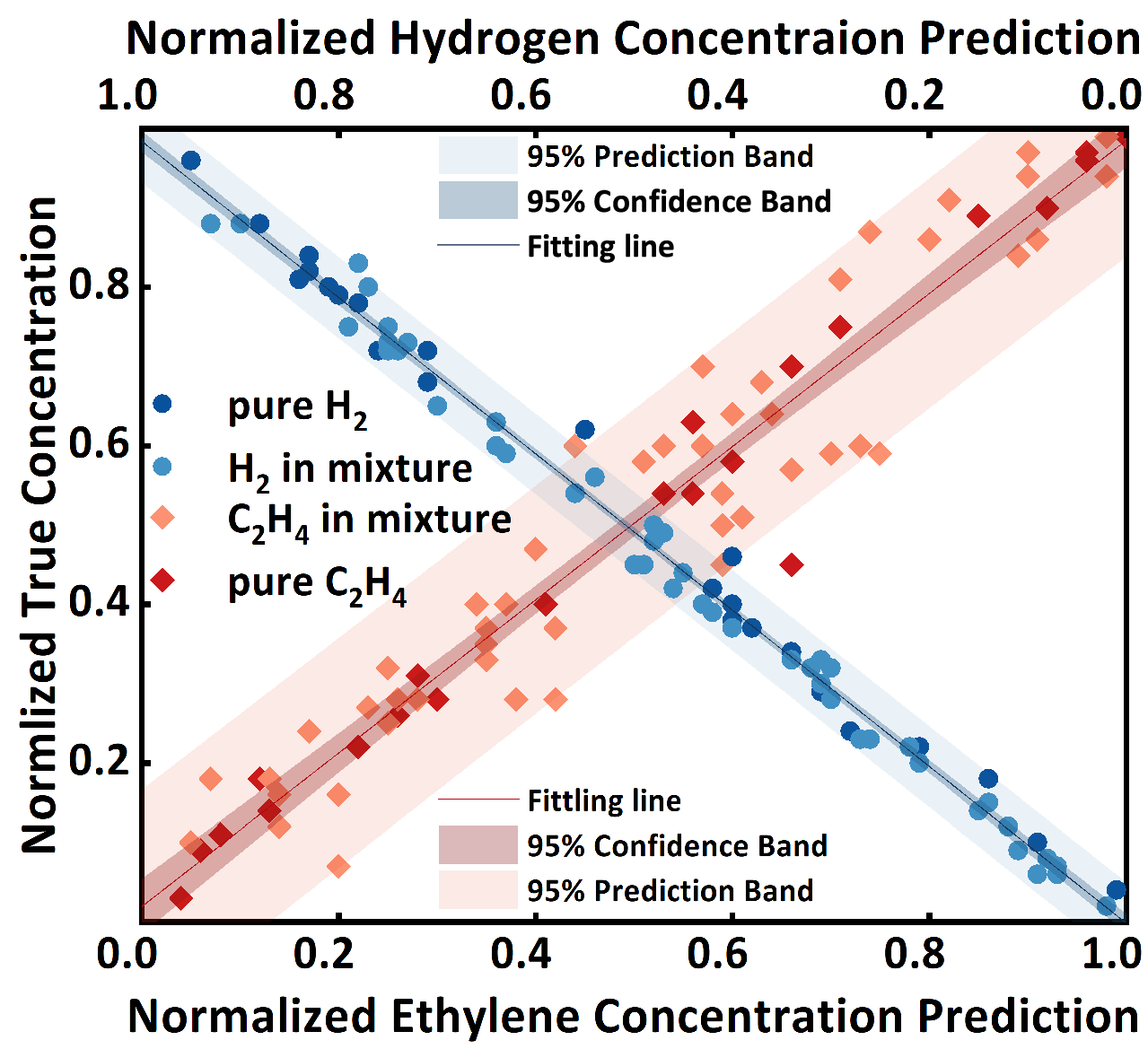}
      \end{minipage}}
      \subfigure[LSTMA/H\textsubscript{2}-C\textsubscript{2}H\textsubscript{4}]
      {\begin{minipage}[c]{0.24\textwidth}
	 \centering
      \includegraphics[width=\textwidth]{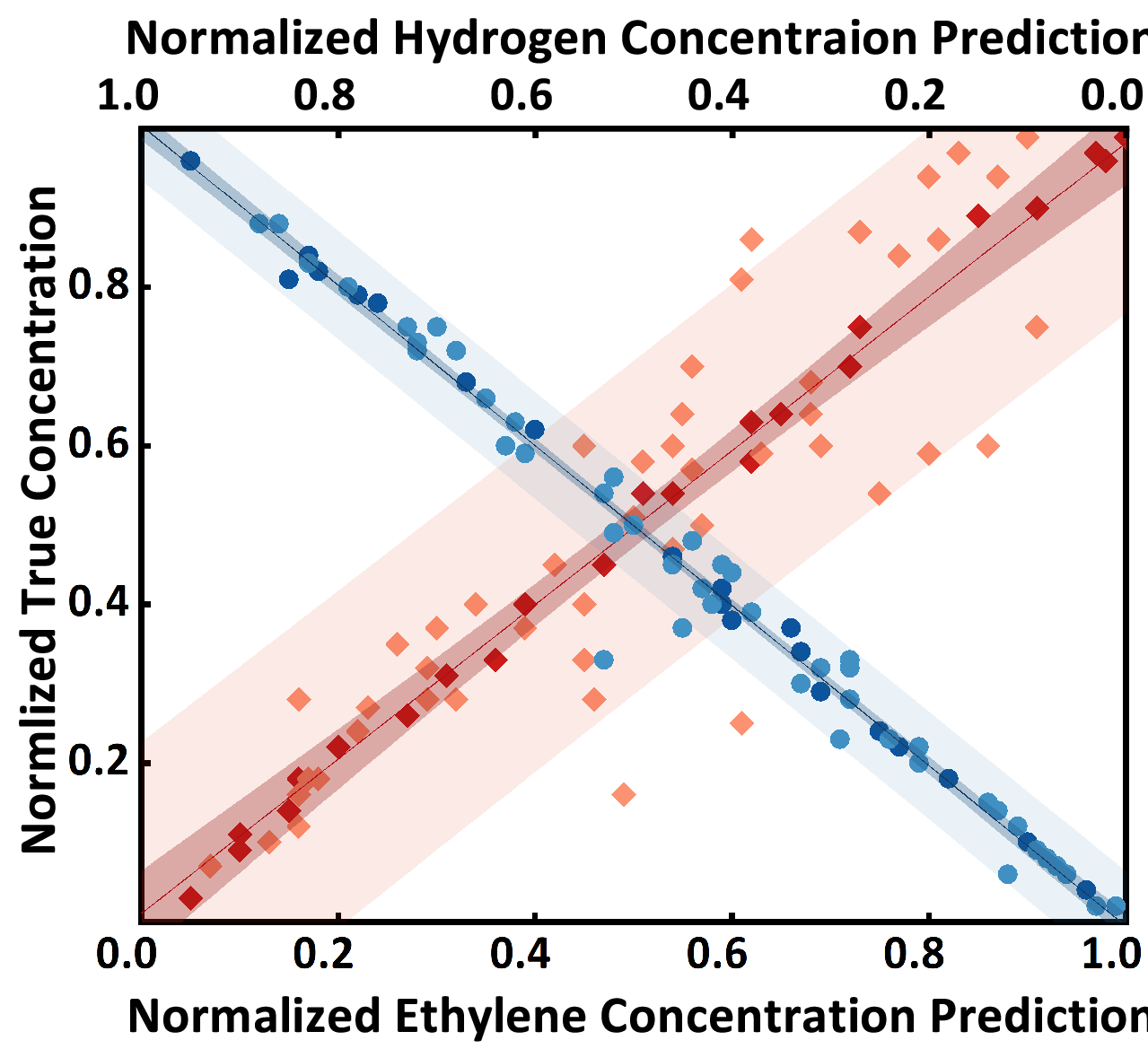}
      \end{minipage}}
      \subfigure[2TCN/H\textsubscript{2}-C\textsubscript{2}H\textsubscript{4}]
      {\begin{minipage}[c]{0.24\textwidth}
	 \centering
      \includegraphics[width=\textwidth]{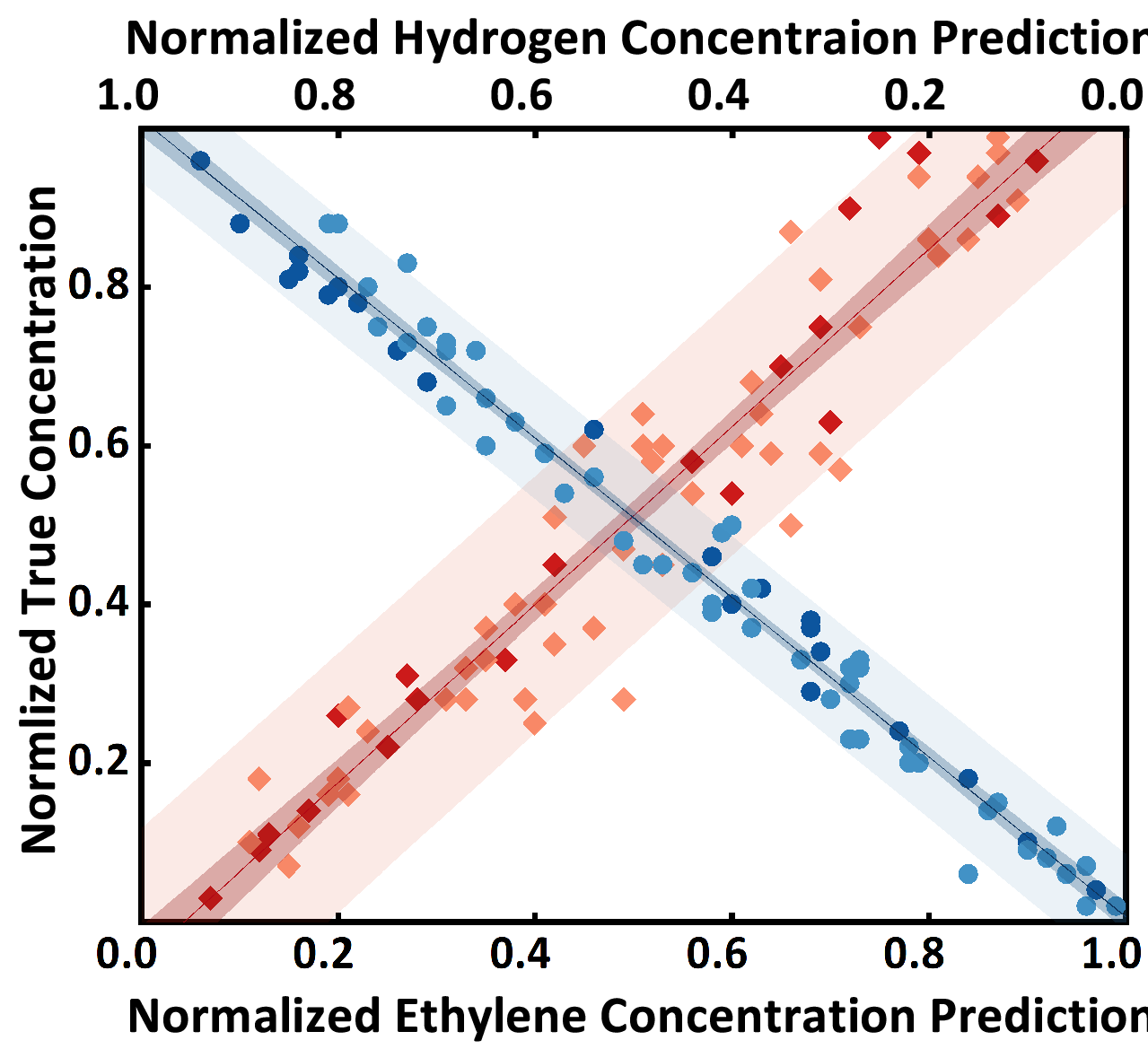}
      \end{minipage}}
      \subfigure[ViT/H\textsubscript{2}-C\textsubscript{2}H\textsubscript{4}]
      {\begin{minipage}[c]{0.24\textwidth}
	 \centering
      \includegraphics[width=\textwidth]{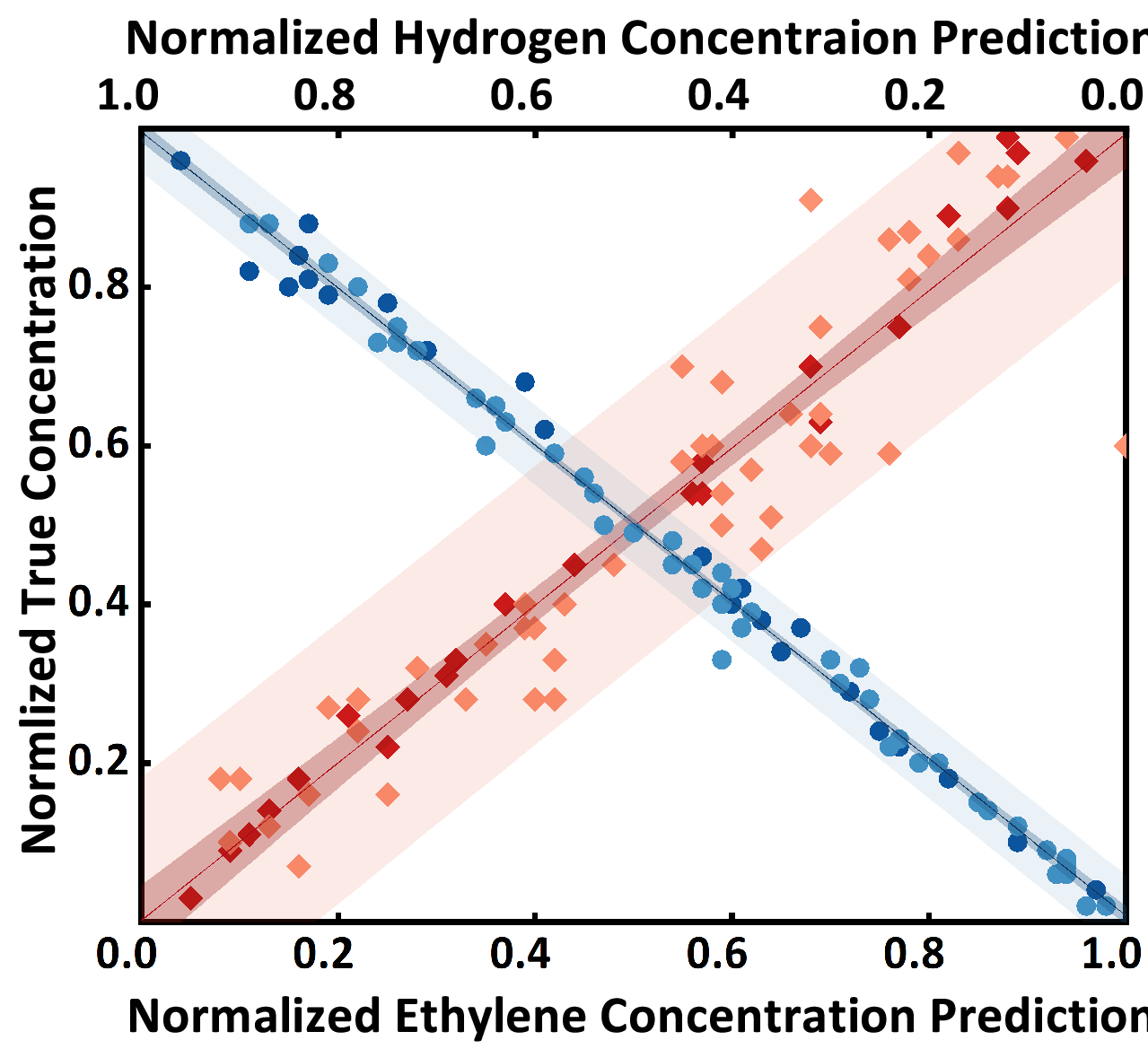}
      \end{minipage}}
\caption{The concentration estimation error of GraphANet, LSTMA, 2TCN, and ViT in three groups, CO-C\textsubscript{2}H\textsubscript{4}, CH\textsubscript{4}-C\textsubscript{2}H\textsubscript{4}, and H\textsubscript{2}-C\textsubscript{2}H\textsubscript{4}.}
\label{ed}
\end{figure*}

\begin{figure}[!ht]
\centering
	\subfigure[UCI Group A]
	{\begin{minipage}[c]{0.9\columnwidth}
      \centering
      \includegraphics[width=\columnwidth]{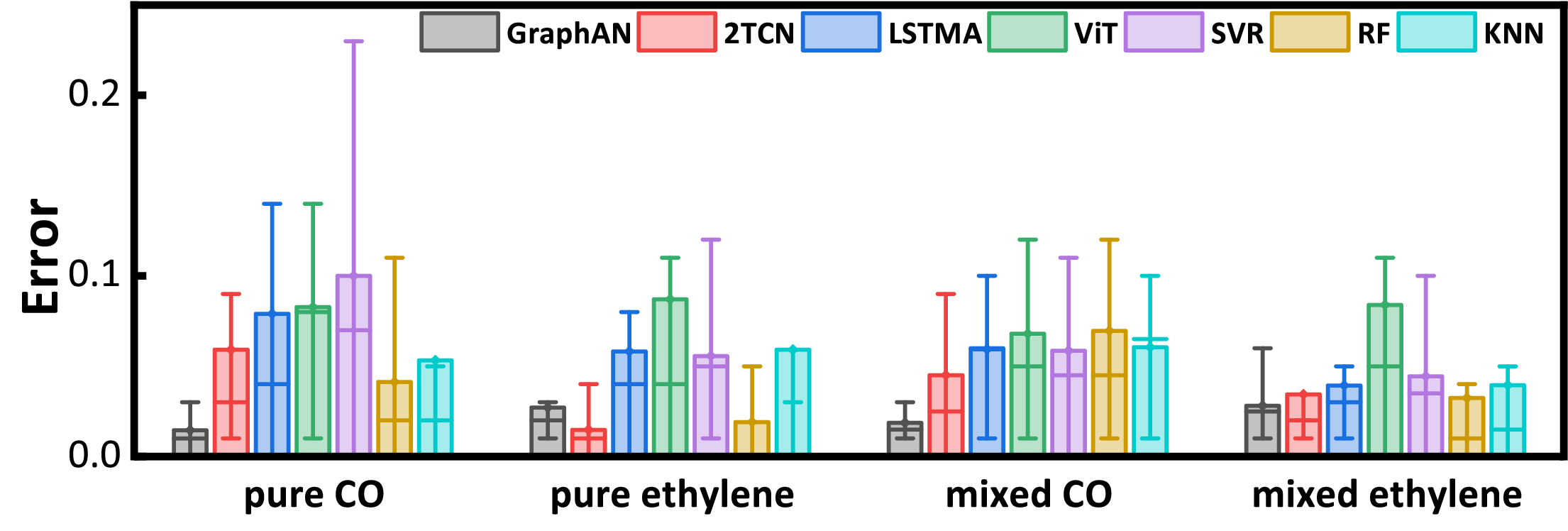}
      \end{minipage}}
	\subfigure[UCI Group B]
	{\begin{minipage}[c]{0.9\columnwidth}
	 \centering
      \includegraphics[width=\columnwidth]{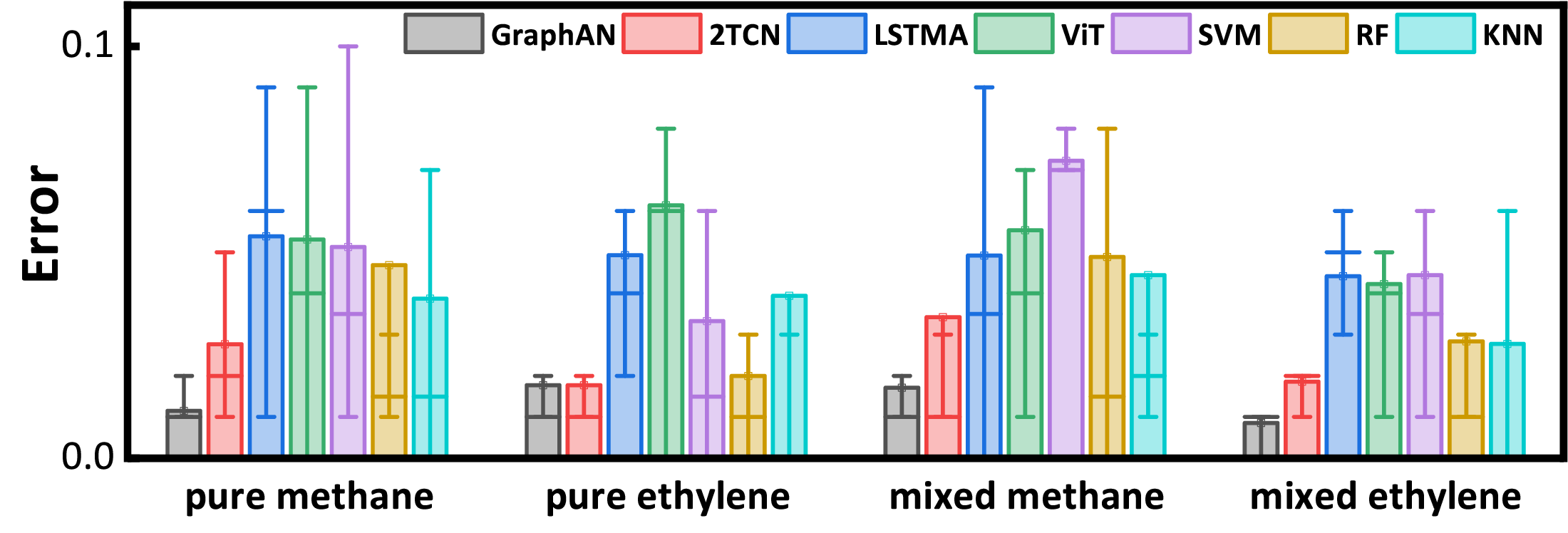}
      \end{minipage}}
	\subfigure[Custom Dataset]
      {\begin{minipage}[c]{0.9\columnwidth}
	 \centering
      \includegraphics[width=\columnwidth]{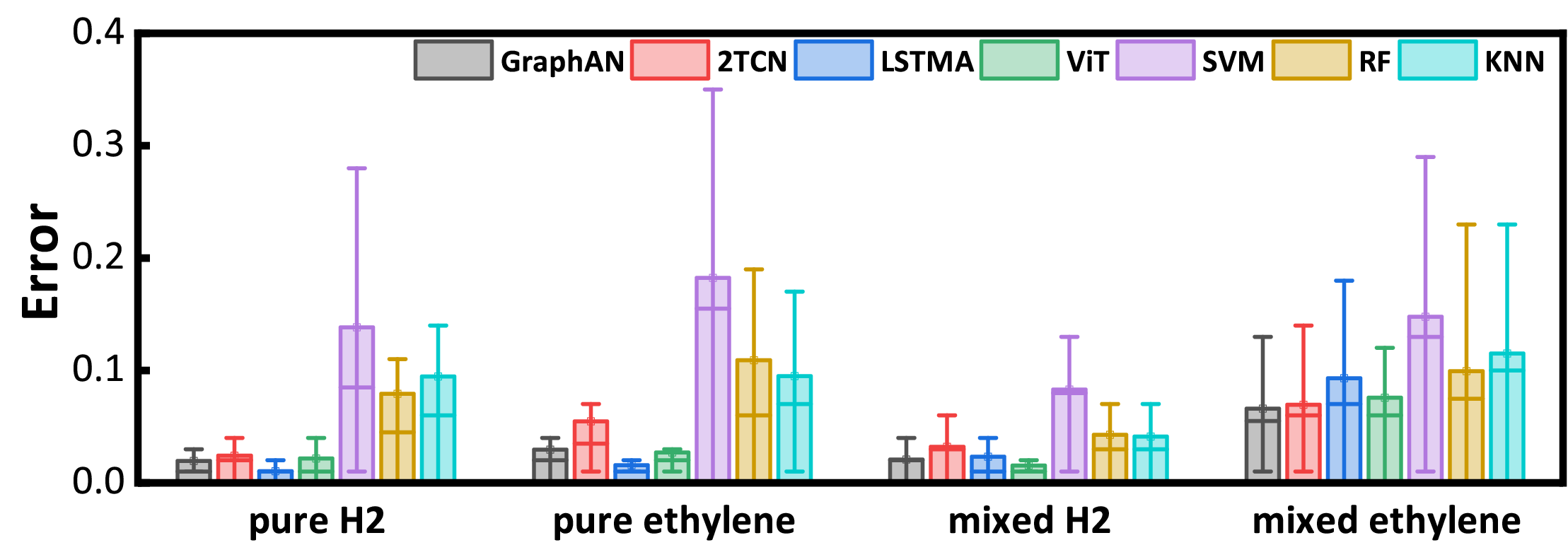}
      \end{minipage}}
\caption{The error distribution of different gas components in different groups.}
\label{cep}
\end{figure}

As presented in Table \ref{tab:class performance}, the proposed GraphCapsNet model demonstrated exceptional performance on both the UCI and custom datasets, achieving high accuracy rates of 0.983 and 0.990, respectively. In comparison, the two deep learning models, PSCFormer and TeTCN, showed slightly lower performance on the UCI dataset, with accuracy rates of 0.958 and 0.832, respectively. On the custom dataset, PSCFormer achieved an accuracy of 0.979, while TeTCN performed slightly lower at 0.969. Overall, their weighted accuracies were also lower than those of GraphCapsNet, with PSCFormer attaining 0.967 and TeTCN 0.888. Meanwhile, the three traditional machine learning models demonstrated relatively consistent performance across both datasets but failed to match the accuracy levels of GraphCapsNet.

\subsection{Classification results analysis}
Fig. \ref{result} illustrates the training, validation, and test accuracies and losses for the models on both the UCI and custom datasets. On the UCI dataset, the accuracy curve shows a sharp increase during the initial epochs, followed by a plateau. The test accuracy is slightly lower than the training and validation accuracies, suggesting strong generalization with minimal overfitting. Similarly, the loss curve exhibits a rapid decline in the initial epochs before stabilizing, indicating efficient learning and good convergence. The close alignment of test loss with training and validation losses further supports the model's robust generalization capabilities. Comparable trends are observed in the accuracy and loss curves for the custom dataset, demonstrating that the model performs well and generalizes effectively on this dataset.

As shown in Fig. \ref{figcm}, a comparison of the confusion matrices for the GraphCapsNet, PSCFormer, and TeTCN models reveals that GraphCapsNet achieves the highest classification accuracy. GraphCapsNet demonstrates strong diagonal dominance, indicating more accurate predictions across various gases and fewer misclassifications, as evidenced by lower off-diagonal values. While PSCFormer performs reasonably well, it exhibits more noticeable misclassifications, particularly in distinguishing between CH\textsubscript{4} and CO-C\textsubscript{2}H\textsubscript{4}, as well as between CO and CO-C\textsubscript{2}H\textsubscript{4}. Additionally, isolated cases of misclassification occur, such as Air-C\textsubscript{2}H\textsubscript{4} being predicted as C\textsubscript{2}H\textsubscript{4} and CH\textsubscript{4} being identified as air. TeTCN shows the weakest performance, with higher confusion rates, particularly between CO and CO-C\textsubscript{2}H\textsubscript{4}, and C\textsubscript{2}H\textsubscript{4} and air. These results clearly demonstrate GraphCapsNet's superior performance in gas recognition tasks, achieving higher accuracy and stronger generalization compared to PSCFormer and TeTCN.


\subsection{Overview of concentration estimation results}

This study evaluated the performance of concentration estimation models across three datasets: UCI group A, UCI group B, and a custom dataset. As shown in Table \ref{rrt}, the GraphANet model consistently outperformed other models, achieving the highest $\rm R^2$
values and the lowest MAE, RMSE, and MAPE across all groups. These results demonstrate GraphANet's exceptional accuracy and reliability in concentration estimation tasks.\par

Other models produced varying levels of performance depending on the group. The RF model performed well in group A, achieving an $\rm R^2$ of 0.839 and an MAE of 0.040, but overall, it fell short compared to GraphANet. The 2TCN and LSTMA models showed strong performance but were slightly less effective than GraphANet. In contrast, the ViT model exhibited more variability, with lower $\rm R^2$ values and higher MAPE in certain cases. These findings underscore the advanced capabilities of deep learning models, particularly GraphANet, in delivering high-precision concentration estimations, setting a benchmark for future research in this domain.

\subsection{Concentration estimation results analysis}
To further analyze these results, we created error charts for the gas components in each group, as illustrated in Fig. \ref{ed}, and error distribution charts for different components across various groups, as shown in Fig. \ref{cep}. 

Fig. \ref{cep} illustrates the overview of the error distribution of various models across three groups. Notably, the GraphANet model consistently achieves the lowest error in most scenarios compared to other models and demonstrates superior performance with more stable and lower error distributions. GraphANet excels in scenarios involving pure and mixed hydrogen, particularly in the custom dataset, highlighting its robustness and generalization capabilities in complex gas detection tasks.

In Fig. \ref{ed}, the distribution of errors can be observed in more detail. From a horizontal perspective, we can observe that the GraphANet model demonstrates its advantages compared to other models, with narrower confidence and prediction intervals. Although some models achieve high accuracy in estimating the concentrations of different gases within specific datasets, their results often exhibit an unbalanced distribution, usually at the expense of accuracy for another gas in the same group. For instance, the LSTMA model performs well in identifying H\textsubscript{2} components in the custom dataset, but its performance in recognizing C\textsubscript{2}H\textsubscript{4} components is significantly lower than that of other models.

A vertical comparison of the error charts in Fig. \ref{ed} reveals the inherent limitations of gas mixture concentration estimation models. Across all models, the CH\textsubscript{4}-C\textsubscript{2}H\textsubscript{4} group consistently achieves better results. This is likely because the CH\textsubscript{4}-C\textsubscript{2}H\textsubscript{4} pair exhibits more similar reductive properties compared to the CO-C\textsubscript{2}H\textsubscript{4} and H\textsubscript{2}-C\textsubscript{2}H\textsubscript{4} groups. When using metal-oxide-semiconductor sensors, gas components with similar reductive or oxidative properties tend to yield more balanced results. Among these gas groups, H\textsubscript{2} demonstrates the strongest reductive properties, which explains why H\textsubscript{2} recognition in the custom dataset consistently dominates across different models. However, this discrepancy is largely attributed to hardware system limitations, highlighting the need for a more generalized algorithm as a future research direction.


In addition, an ablation study was conducted to evaluate the impact of the graph module in GraphANet. As shown in Table \ref{rrt}, the ViT model performs comparably to other deep learning models on the custom dataset and UCI Group B. Notably, for the H\textsubscript{2} component in the custom dataset, ViT achieves better concentration estimation results than GraphANet. However, significant inconsistencies emerge in UCI Group A. A closer examination of UCI Group A reveals that the $\rm R^2$ values for mixed gas detection are low and even negative, indicating that ViT’s performance on mixed gases is adversely affected by the inclusion of pure samples, which obscure its inability to model complex mixtures accurately. The absence of a dedicated feature extraction component in ViT often leads to unstable performance—yielding strong results for H\textsubscript{2} detection but weaker performance for CO and C\textsubscript{2}H\textsubscript{4}. For the CH\textsubscript{4}-C\textsubscript{2}H\textsubscript{4} group, due to the relatively concentrated data distribution, ViT performs comparably to other models.
\section{Conclusion}

This study successfully tackled the challenge of developing deep learning models for gas mixture identification that generalize across heterogeneous datasets. Our proposed models, GraphCapsNet and GraphANet, consistently performed on diverse datasets without requiring retraining for each new dataset, a limitation in current methods. This highlights the models' adaptability and scalability, which are crucial for real-world applications. Although validation was performed on specific datasets under controlled lab conditions, the results indicate significant potential for broader applications. Further research is required to test robustness in more varied, real-world situations and explore extensions to other sensor types.

By introducing adaptable and scalable models, this research contributes to developing more versatile gas mixture identification systems. These findings offer practical solutions for industries needing real-time, accurate gas monitoring. Future efforts will aim to refine these models for large-scale deployment in industrial settings.

\section*{Acknowledgments}

We are grateful for the efforts of our colleagues from the Sino German Center of Intelligent Systems in the College of Electronic and Information Engineering at Tongji University.










\end{document}